\pdfoutput=1

\documentclass[11pt]{article}

\usepackage[preprint]{acl}

\usepackage{times}
\usepackage{latexsym}

\usepackage[T1]{fontenc}

\usepackage[utf8]{inputenc}

\usepackage{microtype}

\usepackage{inconsolata}

\usepackage{graphicx}

\usepackage{makecell}
\usepackage{algorithmicx}
\usepackage{algorithm}
\usepackage{algpseudocode}
\usepackage{times}
\usepackage{latexsym}
\usepackage{xcolor}
\usepackage{soul}
\usepackage{microtype}
\usepackage{inconsolata}
\usepackage{graphicx}
\usepackage{enumitem}
\usepackage{array}
\usepackage{url}
\usepackage{booktabs}
\usepackage{amssymb}
\usepackage{multirow}
\usepackage{pifont}
\usepackage{courier}
\usepackage{lipsum}
\usepackage{cleveref}
\usepackage{catchfile}
\usepackage[defaultcolor=magenta]{changes}
\usepackage{multicol}
\usepackage{array}
\usepackage{booktabs} 
\usepackage{catchfile} 
\usepackage{lipsum}
\usepackage{listings}
\usepackage{xcolor}
\usepackage{listingsutf8}

\usepackage[T1]{fontenc}  
\usepackage[utf8]{inputenc}
\usepackage{kotex}

\title{KorMedMCQA: Multi-Choice Question Answering Benchmark for \ Korean Healthcare Professional Licensing Examinations}

\author{Sunjun Kweon$^{1}$$^{*}$, Byungjin Choi$^{2}$\thanks{\hspace{0.05cm} Equal contribution}, Gyouk Chu$^{1}$, Junyeong Song$^{3}$, Daeun Hyeon$^{2}$\\
    \textbf{Sujin Gan$^{2}$, Jueon Kim$^{4}$, Minkyu Kim$^{2}$, Raewoong Park$^{2}$, Edward Choi$^{1}$}\\
  KAIST$^{1}$ Ajou University School of Medicine$^{2}$ 
  \\ UNIST$^{3}$ Kyunghee University College of Dentistry$^{4}$
  }

\begin{document}
\maketitle
\begin{abstract}
We present KorMedMCQA, the first Korean Medical Multiple-Choice Question Answering benchmark, derived from professional healthcare licensing examinations conducted in Korea between 2012 and 2024.
The dataset contains 7,469 questions from examinations for doctor, nurse, pharmacist, and dentist, covering a wide range of medical disciplines.
We evaluate the performance of 59 large language models, spanning proprietary and open-source models, multilingual and Korean-specialized models, and those fine-tuned for clinical applications.  
Our results show that applying Chain of Thought (CoT) reasoning can enhance the model performance by up to 4.5\% compared to direct answering approaches.
We also investigate whether MedQA, one of the most widely used medical benchmarks derived from the U.S. Medical Licensing Examination, can serve as a reliable proxy for evaluating model performance in other regions—in this case, Korea.
Our correlation analysis between model scores on KorMedMCQA and MedQA reveals that these two benchmarks align no better than benchmarks from entirely different domains (\textit{e.g.,} MedQA and MMLU-Pro).
This finding underscores the substantial linguistic and clinical differences between Korean and U.S. medical contexts, reinforcing the need for region-specific medical QA benchmarks.
To support ongoing research in Korean healthcare AI, we publicly release the KorMedMCQA via Huggingface \url{https://huggingface.co/datasets/sean0042/KorMedMCQA}.
\end{abstract}

    \begin{figure}[t]
    \includegraphics[width=0.48\textwidth,trim={10 10 10 10}, clip]{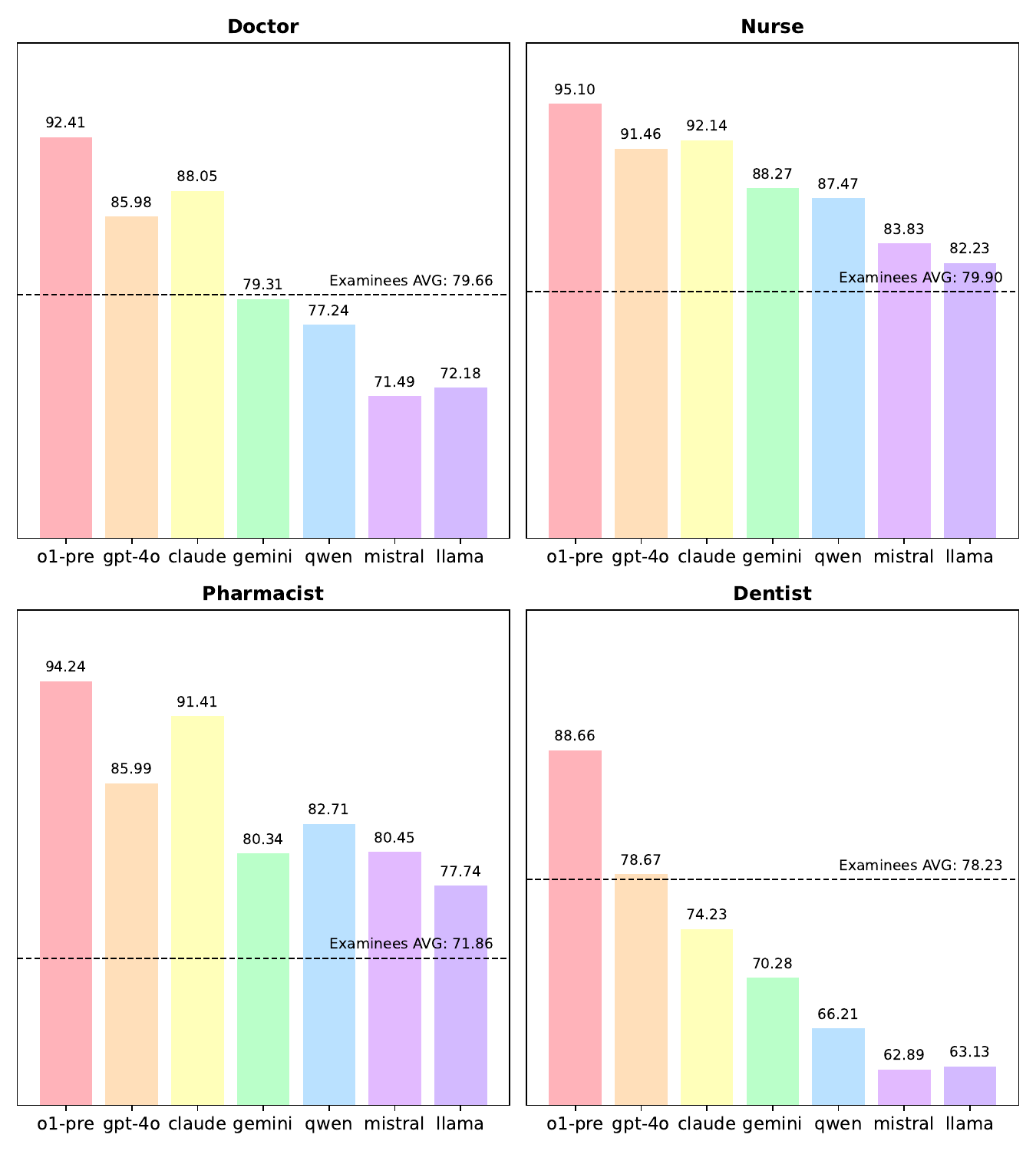}
    \caption{Evaluation results for the Doctor, Nurse, Pharmacist, and Dentist subsets of KorMedMCQA across selected LLMs. The dotted line represents the average scores of actual examinees from official exam results.}
    \vspace{-3mm}
    \label{fig1}
    \end{figure}

\section{Introduction}

Assessing a language model’s medical expertise often involves evaluating its performance on professional medical licensing exam questions \citep{nori2023capabilities, nori2023can, chen2023meditron, toma2023clinical, han2023medalpaca, singhal2023large,singhal2023towards,dhakal2024gpt}. 
This approach assesses not only the model’s clinical knowledge but also its potential utility in real-world healthcare scenarios. 
The MedQA \citep{jin2021disease} dataset, derived from the U.S. Medical Licensing Examination (USMLE), has been a cornerstone in such evaluations and remains a primary resource for testing model performance in healthcare contexts.
Yet an important question remains: \textit{Can a benchmark rooted in U.S.-based medical exams reliably evaluate the performance of language models intended for other countries' healthcare systems?}

In the case of Korea, the answer is likely no, for three reasons.
First, a model's effectiveness depends on its understanding of the language.
Since even identical questions translated across languages can pose entirely different challenges \citep{hu2020xtreme}, a model's performance in different medical languages remains uncertain without proper testing.
Second, treatment guidelines can vary significantly across different national backgrounds.
For instance, guidelines for acute pyelonephritis (APN) differ between the U.S. \citep{colgan2011diagnosis} and Korea \citep{korean2018antibiotic}, which affects how medical questions should be answered.
Lastly, medical exams assess not only clinical knowledge but also an understanding of relevant laws and regulations, which vary by country.
In conclusion, excelling on a U.S.-based exam does not guarantee success on a Korean exam due to linguistic, national, and regulatory differences.


To address these gaps, we introduce KorMedMCQA, the first Korean Medical Multi-Choice Question Answering dataset derived from licensing examinations for doctors, nurses, pharmacists, and dentists in South Korea (see Figure \ref{appendix:figure1} in Appendix for examples).
This dataset, comprising 7,469 questions from exams administered between 2012 and 2024, offers a comprehensive tool for evaluating the model capability to understand and reason over medical knowledge within the Korean context.

We evaluated 59 Large Language Models (LLMs)—both proprietary and open-source, multilingual and Korean-specialized, and clinically adapted—using KorMedMCQA.
Among proprietary models, o1-preview \citep{openai2024b} scored highest average score (\textit{92.72}), while Qwen2.5-72B \citep{yang2024qwen2} led open-source models (\textit{78.86}) (see each subject scores in Figure \ref{fig1}).
To further enhance model performance, we applied Chain of Thought (CoT) reasoning \citep{wei2022chain}, which improved scores up to a 4.5\% compared to direct answering.
Building upon CoT outputs, we conducted an error analysis by medical professionals, identifying areas for improvement.
Finally, to experimentally confirm the necessity of country-specific medical benchmarks, we examined correlation between model scores on KorMedMCQA and MedQA. 
Comparisons with correlations from benchmarks of unrelated domains further highlighted the necessity of country-specific datasets for evaluating real-world applicability in medical context.

In order to invite further exploration in Korean healthcare AI, we release our data publicly on HuggingFace \citep{wolf2019huggingface} and provide a evaluation script via LM-Harness \citep{eval-harness}.

\section{Related Works}

Several prior benchmarks have been developed using medical licensing exams.
Among them, MedQA \citep{jin2021disease}, which is based on the U.S. medical exam, is the most commonly used and serves as a critical standard for assessing medical language models.
Other country-specific benchmarks have also been proposed, including HeadQA (Spain) \citep{vilares2019head}, FrenchMCQA (France) \citep{labrak2023frenchmedmcqa}, MedMCQA (India) \citep{pal2022medmcqa}, and CMExam (China) \citep{liu2024benchmarking}.
While each dataset reflects the distinct clinical context of its respective country, our study is the first to directly compare model performance on MedQA and another regional benchmark (KorMedMCQA), providing empirical evidence that highlights the importance of region-specific evaluation resources.

\section{Data}

\subsection{Data Collection}

We collected questions from Korean medical licensing examinations, including exams for doctors (2012–2024), nurses and pharmacists (2019–2024), and dentists (2020–2024).
All questions and answers were sourced from official releases by the Korea Health Personnel Licensing Examination Institute\footnote{\url{https://www.kuksiwon.or.kr}}, which made these exams publicly available for academic use.
The data collection involved crawling PDFs of exam materials and organizing the content into structured datasets.
During the data preprocessing, we excluded about 100 questions with multiple correct answers to maintain consistency. Additionally, nearly 1,000 questions requiring image or video contents were removed, as these elements are often masked in the original PDFs due to copyright issues.
Each dataset entry includes a unique identifier (denoting the year, session, and question number), the question text, five answer choices, and the index of the correct answer index.

\subsection{Data Statistics}

The dataset consists of 7,469 multiple-choice questions: 2,489 for doctors, 1,751 for nurses, 1,817 for pharmacists, and 1,412 for dentists.
Table \ref{appendix:split_number} details the yearly distribution and how they were split into training (pre-2021), development (2021), and test sets (2022–2024).
The dataset cover a broad range of medical disciplines—3 major subjects for doctors (e.g., General Medicine), 7 for nurses (e.g., Pediatric Nursing), 21 for pharmacists (e.g., Biochemistry), and 13 for dentists (e.g., Orthodontics). 
A detailed breakdown of the subjects is provided in Table \ref{appendix:subjects} provides a detailed subject breakdown.

\section{Experimental Setup}

\subsection{Evaluation Method}

We evaluated various LLMs on the KorMedMCQA's test set using a 5-shot setting, with shots selected from the development set.
Generated model outputs were processed using regular expressions to extract predicted answers, which were then compared to the correct answer indices.
For proprietary models (\textit{e.g.,} GPT), outputs were generated directly using each model's API.
Open-source models were evaluated using LM-Harness \citep{eval-harness}.
A more detailed description of the evaluation method can be found in Appendix \ref{appendix:evaluation}.

\subsection{Models}

We evaluated 59 LLMs, comprising 13 proprietary models and 46 open-source models.
For proprietary models, we covered all available versions of GPT, Claude, and Gemini.
The open-source models included multilingual pretrained LLMs such as LLaMA3 \citep{dubey2024llama}, Mistral \citep{mistral2024}, Qwen \citep{yang2024qwen2}, Gemma \citep{team2024gemmab}, Yi \citep{young2024yi}, and SOLAR \citep{kim2023solar}.
Additionally, we evaluated models specifically trained for Korean including EXAONE \citep{an2024exaone} (pretrained from scratch) and EEVE \citep{kim2024efficient} (adapted from SOLAR), as well as English models trained for clinical context such as Meditron \citep{chen2023meditron} (adapted from LLaMA-3.1) and Meerkat \citep{kim2024small} (adapted from LLaMA-3).

    \begin{table}[t]
    \centering
    \setlength{\heavyrulewidth}{1.0pt}
    \resizebox{1.0\columnwidth}{!}{
        \newcommand{\mytable}{\begin{tabular}{lcccccc}
\toprule
\multicolumn{1}{c}{\textbf{Model}} & \multicolumn{1}{c}{\textbf{Doctor}} & \multicolumn{1}{c}{\textbf{Nurse}} & \multicolumn{1}{c}{\textbf{Pharm}} & \multicolumn{1}{c}{\textbf{Dentist}} & \multicolumn{1}{c}{\textbf{Avg}} & \multicolumn{1}{c}{\textbf{MedQA}} \\ 
\midrule
\multicolumn{7}{c}{\textbf{Human Performance}}  \\ \midrule
Examinees Average&79.66&79.90&71.86&78.23&77.05&- \\
\midrule 

\multicolumn{7}{c}{\textbf{Proprietary Models}}                                                                                                                                       \\ \midrule
\textbf{o1-preview} & 92.41 & 95.10 & 94.24 & 88.66 & \textbf{92.72} & 94.19 \\
o1-mini & 88.28 & 85.88 & 87.68 & 69.91 & 82.45 & 90.18 \\
gpt-4o & 85.98 & 91.46 & 85.99 & 78.67 & 85.61 & 86.96 \\
gpt-4o-mini & 63.22 & 83.14 & 71.30 & 59.06 & 70.29 & 72.90 \\
claude-3-5-sonnet&88.05&92.14&91.41&74.23&86.51 & 82.95 \\
claude-3-sonnet&66.90&78.13&63.73&43.25&62.87 & 65.91 \\
gemini-1.5-pro&79.31&88.27&80.34&70.28&79.79 & 76.67 \\
gemini-1.5-flash&68.51&82.46&74.92&59.68&72.09 & 67.09 \\
\midrule

\multicolumn{7}{c}{\textbf{Multilingual Models}}                                                                                                                    \\ \midrule
Llama-3.1-70B-Instruct&72.18&82.23&77.74&63.13&74.31&79.42 \\
Llama-3.1-8B-Instruct&40.69&59.57&55.25&42.05&50.85&63.32 \\
Llama-3-70B-Instruct&66.67&81.21&73.90&63.01&72.05&75.88 \\
Llama-3-8B-Instruct&38.16&55.92&48.14&41.68&47.23&62.37 \\
\textbf{Qwen2.5-72B-Instruct}&77.24&87.47&82.71&66.21&\textbf{78.86}&77.93 \\
Qwen2.5-7B-Instruct&51.49&27.33&63.39&39.58&44.73&61.19 \\
Qwen2.5-1.5B-Instruct&33.33&47.72&35.48&32.43&37.92&45.56 \\
Mistral-Large-Instruct&71.49&83.83&80.45&62.89&75.41&79.65 \\
Mistral-Nemo-Instruct&42.30&65.83&58.31&43.03&54.07&60.88 \\
gemma-2-27b-it&62.30&78.36&72.20&58.57&68.89&67.79 \\
gemma-2-9b-it&20.69&61.73&43.95&42.42&45.36&62.53 \\
gemma-2-2b-it&13.33&39.41&27.80&28.11&29.18&42.18 \\
Yi-1.5-34B-Chat&36.09&51.03&48.81&38.84&44.93&63.55 \\
Yi-1.5-9B-Chat&31.26&44.99&38.98&34.77&38.48&52.24 \\
SOLAR-10.7B-Instruct&41.84&58.54&52.77&37.85&48.85&54.52 \\
\midrule
\multicolumn{7}{c}{\textbf{Korean Specific Models}}                                                                                                                  \\ \midrule
EEVE-10.8B&48.97&66.17&58.19&42.42&54.94&53.50 \\
EXAONE-3.0-7.8B&48.28&67.65&56.61&45.87&55.73&50.59 \\

\midrule
\multicolumn{7}{c}{\textbf{Medical Continual Pretrained \& Finetuned Models}}                                                                                                                      \\ \midrule
Meditron3-70B&70.57&82.69&75.82&62.64&73.51&79.65 \\
Meditron3-8B&38.39&58.20&54.92&39.83&49.42&61.98 \\
meerkat-70b-v1.0&69.20&80.75&75.59&60.06&71.99&77.85 \\
meerkat-8b-v1.0&38.62&56.61&45.65&40.57&46.46&62.53

\\ \bottomrule
\end{tabular}}
            \begin{tabular}{l}
\toprule
\textbf{질문:} 45세 여자가 입 주위와 손발이 저리다고 한다. 3시간 전 갑상샘암으로 오른쪽 갑상샘절제를 받았다. 혈압 119/78 mmHg, 맥박 89회/분, 호흡 29회/분, 체온 36.7℃이다. 목은 붓지 않았다. 가슴 청진에서 호흡음은 정상으로 들린다. 혈액검사 결과는 다음과 같다. 조치는?혈액: Na +/K+/Cl- 143/3.8/101 meq/L Ca/P 8.6/4.1 mg/dL, 부갑상샘호르몬(iPTH) 19 pg/mL (참고치, 8～51) 갑상샘자극호르몬 3.21 mIU/L (참고치, 0.34～4.25) 유리 T4 1.3 ng/dL (참고치, 0.8～1.7) 동맥혈기체(대기호흡) : pH 7.49, PaCO2 29 mmHg PaO2 97 mmHg, HCO3- 22.2 meq/L \\
\textbf{보기:} A. 경과관찰 B. 산소흡입 C. 재호흡주머니 적용 D. 레보타이록신 투여 E. 10\% 글루콘산칼슘 정맥주사 \\
\textbf{정답:} C \\
\textbf{GPT:} 차근차근 생각해봅시다. 45세 여자는 갑상샘암으로 인해 오른쪽 갑상샘절제를 받은 후 입 주위와 손발이 저리는 증상을 호소하고 있습니다. 이러한 증상은 저칼슘혈증과 관련이 있을 수 있습니다. 혈액검사 결과를 보면, 혈청 칼슘 수치가 8.6 mg/dL로 정상 하한선에 가깝지만, 수술 후 저칼슘혈증의 증상으로 나타날 수 있습니다. 또한, 부갑상샘호르몬(iPTH) 수치가 정상 범위 내에 있지만, 수술 후 부갑상샘 기능 저하로 인해 상대적으로 낮을 수 있습니다. 동맥혈기체 분석에서 pH가 7.49로 알칼리증을 보이고 있으며, 이는 호흡성 알칼리증을 시사합니다. 그러나, 저칼슘혈증으로 인한 신경과민 증상이 더 우선적으로 해결되어야 합니다. 이러한 경우, 10\% 글루콘산칼슘 정맥주사를 통해 칼슘을 보충하는 것이 적절한 조치입니다. 따라서, 정답은 (E) 10\% 글루콘산칼슘 정맥주사입니다. \\
\textbf{Claude:} 차근차근 생각해봅시다.
1. 임상적 판단:
- 갑상샘 수술 후 발생한 저린 증상은 수술 중 부갑상샘 손상이나 일시적인 기능저하로 인한 저칼슘혈증을 시사
- 호흡이 빨라진 것은 저칼슘혈증으로 인한 보상기전일 수 있음
- 입 주위 저림은 Chvostek 징후일 수 있으며, 이는 저칼슘혈증의 전형적인 증상
2. 필요한 조치:
- 급성 저칼슘혈증에 대한 즉각적인 교정이 필요
- 10\% 글루콘산칼슘의 정맥 투여가 가장 적절한 치료
따라서, 정답은 (E) 입니다. \\
\textbf{Gemini:} 차근차근 생각해봅시다. 45세 여자 환자는 갑상샘암으로 오른쪽 갑상샘절제술을 받은 후 입 주위와 손발 저림을 호소합니다. 이는 부갑상샘 기능저하증을 시사하는 증상입니다. 혈액 검사 결과를 보면 칼슘 수치는 8.6 mg/dL로 정상 범위보다 약간 낮고, 부갑상샘호르몬(iPTH) 수치는 19 pg/mL로 정상 범위 안에 있지만, 갑상샘절제술 후에는 일반적으로 iPTH 수치가 상승해야 합니다. 이는 수술 중 부갑상샘이 손상되었을 가능성을 시사하며, 환자의 저칼슘혈증 증상과 함께 부갑상샘 기능저하증을 의심하게 합니다. 따라서 저칼슘혈증 증상을 조절하기 위해 10\% 글루콘산칼슘 정맥주사를 투여하는 것이 적절한 조치입니다. 따라서, 정답은 (E) 입니다.\\
\textbf{설명: } 이 환자는 얼핏 갑상샘 수술 이후 부갑상샘 저하증으로 생각될 수 있습니다. 각 모델은 모두 부갑상샘 기능저하를 의심하고 있습니다. 이에 따라, 각 모델들은 치료로 칼슘 글루코네이트 주사를 추천합니다. 하지만, 일반적으로 칼슘 글루코네이트는 훨씬 더 중증의 저칼슘 혈증에서 사용됩니다.(7.2 mg/dL 이하). 환자는 비교적 경증의 증상만 보이며 칼슘 농도도 Lower boundary 수준입니다. 그에 따라, 굳이 칼슘 글루코네이트 주사까진 필요 없어 보입니다. 또한,환자는 동맥혈 가스검사에서 pH 7.49, PaCO2 29 mmHg로 호흡성 알칼리증을 보이고 분당 29회의 빠른 호흡을 보이고 있습니다. 이를 고려하였을 때 재호흡 주머니를 적용하여 호흡성 알칼리증을 개선하는 것이 필요합니다.  \\
\break
\textbf{Question:} A 45-year-old woman complains of numbness around her mouth and in her hands and feet. She underwent a right thyroid lobectomy for thyroid cancer 3 hours ago. Her vital signs are as follows: blood pressure 119/78 mmHg, pulse rate 89 bpm, respiratory rate 29 breaths/min, and temperature 36.7℃. Her neck is not swollen, and chest auscultation reveals normal breath sounds. Laboratory results are as follows: Blood: Na+/K+/Cl-: 143/3.8/101 mEq/L, Ca/P: 8.6/4.1 mg/dL, parathyroid hormone (iPTH): 19 pg/mL (reference range: 8–51), thyroid-stimulating hormone: 3.21 mIU/L (reference range: 0.34–4.25), free T4: 1.3 ng/dL (reference range: 0.8–1.7). Arterial blood gas (room air): pH 7.49, PaCO2 29 mmHg, PaO2 97 mmHg, HCO3- 22.2 mEq/L. What is the appropriate management? \\ \textbf{Options:} A. Observation B. Oxygen inhalation C. Application of a rebreathing bag D. Administration of levothyroxine E. Intravenous injection of 10\% calcium gluconate \\ \textbf{Answer:} C \\
\textbf{GPT:} Let’s think step by step. A 45-year-old woman complains of tingling around the mouth and in her hands and feet after undergoing a right thyroidectomy due to thyroid cancer. These symptoms could be related to hypocalcemia. From the blood test results, her serum calcium level is 8.6 mg/dL, which is near the lower limit of normal but could manifest as postoperative hypocalcemia. Additionally, her parathyroid hormone (iPTH) levels are within the normal range but may be relatively low due to decreased parathyroid function post-surgery. Arterial blood gas analysis shows a pH of 7.49, indicating alkalosis, which suggests respiratory alkalosis. However, symptoms of neuromuscular irritability due to hypocalcemia should be addressed first. In such cases, administering intravenous 10\% calcium gluconate is an appropriate intervention. Therefore, the correct answer is (E) intravenous 10\% calcium gluconate.  \\
\textbf{Claude:} Let’s think step by step. 3. Clinical judgment: - The tingling symptoms after thyroid surgery suggest hypocalcemia due to potential parathyroid gland damage or temporary dysfunction. - Rapid breathing could be a compensatory mechanism for hypocalcemia. - Tingling around the mouth could indicate Chvostek’s sign, a classic symptom of hypocalcemia. 4. Necessary measures: - Immediate correction of acute hypocalcemia is required. - Intravenous administration of 10\% calcium gluconate is the most appropriate treatment.
Therefore, the correct answer is (E). \\
\textbf{Gemini:} Let’s think step by step. A 45-year-old female patient complains of tingling around the mouth and in her hands and feet following a right thyroidectomy for thyroid cancer. This suggests hypoparathyroidism. Blood test results show a calcium level of 8.6 mg/dL, slightly below normal, and an iPTH level of 19 pg/mL, within the normal range. However, after thyroidectomy, iPTH levels would typically be elevated. This raises the suspicion of parathyroid damage during surgery. Combined with the patient’s hypocalcemic symptoms, hypoparathyroidism is likely. To manage her symptoms, administering intravenous 10\% calcium gluconate is an appropriate measure. Thus, the correct answer is (E).  \\
\textbf{Explanation:} At first glance, this case could be attributed to postoperative hypoparathyroidism following thyroid surgery. All models suspect parathyroid dysfunction and recommend calcium gluconate injections as treatment. However, calcium gluconate is generally used for more severe hypocalcemia (below 7.2 mg/dL). The patient has relatively mild symptoms and a calcium level near the lower boundary of normal. Therefore, intravenous calcium gluconate may not be strictly necessary. Additionally, the patient shows respiratory alkalosis with a pH of 7.49 and PaCO2 of 29 mmHg, along with a rapid respiratory rate of 29 breaths/min. Considering this, applying a rebreathing bag to address respiratory alkalosis would be appropriate.  \\
\bottomrule
\end{tabular}

    }
    \caption{
LLM performance on the KorMedMCQA and MedQA, with top-performing models in KorMedMCQA highlighted in \textbf{bold} for both proprietary and open-source categories. \textit{Avg} denotes weighted average.
    }
    \label{score}
    \end{table}

\section{Results and Analysis}

\subsection{Results} \label{results}
Evaluation results for the selected models, along with average human examinees scores from the licensing exam institute, are summarized in Table \ref{score} (see full results in Appendix \ref{appendix:results}).
A clear trend was observed: models performed best on nurse licensing exams, followed by pharmacist exams, medical doctor exams, and finally dentist exams.
This trend likely reflects the relative scarcity of dental content in the models’ training data compared to medicine, nursing, and pharmacy.

Among proprietary models, the latest version of GPT, o1-preview \citep{openai2024b}, achieved the highest average score (\textit{92.72}), exceeding the average human examinee by 15\%.
Among open-source models, Qwen2.5-72B led with an average score of \textit{78.86}.
Notably, out of all models evaluated, only five proprietary models and one open-source model outperformed the average human examinee, underscoring the need for further improvement.

When including performance on MedQA, additional insights emerged.
First, most models performed better on the English MedQA dataset than on the Korean KorMedMCQA, except for Korean-specific pretrained models such as EEVE and EXAONE. 
Specifically, EEVE, adapted from SOLAR with additional Korean fine-tuning, significantly improved its KorMedMCQA score (from \textit{48.85} to \textit{54.94}), demonstrating even non–medically oriented Korean-specific pretraining can be valuable for tackling Korean medical tasks.

Second, models further trained on English medical context (\textit{e.g.,} Meditron and Meerkat) showed improvements on MedQA but exhibited performance drops on KorMedMCQA.
For example, LLaMA-3.1-70B’s MedQA score increased with Meditron3-70B (\textit{79.42} to \textit{79.65}) and LLaMA-3-70B increased with Meerkat-70B (\textit{75.88} to \textit{77.85}).
However, on KorMedMCQA, Meditron3-70B slightly underperformed the base model (\textit{74.31} to \textit{73.51}), as did Meerkat-70B (\textit{72.05} to \textit{71.99}).
These findings suggest that pretraining / fine-tuning with English medical contents enhances the performance on English datasets (MedQA) but provides limited benefit for Korean tasks (KorMedMCQA).

\subsection{CoT Reasoning \& Error Analysis}

    \begin{table}[t]
    \centering
    \setlength{\heavyrulewidth}{1.0pt}
    \resizebox{1.0\columnwidth}{!}{
        \newcommand{\mytable}{\begin{tabular}{l|ccccc}
\toprule
\textbf{Model} & \textbf{Doctor} & \textbf{Nurse} & \textbf{Pharmacist} & \textbf{Dentist} & \textbf{Average} \\ \midrule
GPT-4o & 85.98 & 91.46 & 85.99 & 78.67 & 85.61 \\
GPT-4o (CoT) & 89.89 & 92.26 & 90.96 & 79.78 & 88.17 \\ \midrule
Claude-3.5-sonnet & 88.05 & 92.14 & 91.41 & 74.23 & 86.51 \\
Claude-3.5-sonnet (CoT) & 92.18 & 93.96 & 93.22 & 84.59 & 90.96 \\ \midrule
Gemini-1.5-pro & 79.31 & 88.27 & 80.34 & 70.28 & 79.79 \\
Gemini-1.5-pro (CoT) & 84.83 & 90.77 & 87.12 & 67.82 & 82.65 \\ \bottomrule
\end{tabular}}
            
    }
    \caption{
LLM performance on KorMedMCQA using CoT Reasoning.
    }
    \label{cot_score}
    \end{table}

We evaluated the top three performing models\footnote{The o1 model was excluded because it internally performs reasoning steps on its own.} on KorMedMCQA to determine whether incorporating Chain-of-Thought (CoT) reasoning \citep{wei2022chain} could further enhance their performance (see Table \ref{cot_score}).
CoT reasoning resulted in noticeable score improvements for all three models: Claude-3.5-sonnet achieved \textit{90.96} (+\textit{4.5}\%), GPT-4o reached \textit{88.17} (+\textit{2.5}\%), and Gemini-1.5-pro scored \textit{82.65} (+\textit{2.9}\%).
Notably, Claude-3.5-sonnet came closest to the performance of the top-ranked o1-preview model, which achieved a score of \textit{92.72}.

To gain deeper insights, we conducted an error analysis of CoT-generated outputs. 
Licensed medical professionals—including a doctor, a nurse, a pharmacist, and a dentist—reviewed errors in 200 randomly selected questions that were incorrectly answered by at least two out of the three models.
In particular, these four professionals each independently reviewed 50 questions and the answers from the three models to provide reason for errors.
The results are summarized below, with details provided in Appendix \ref{appendix_cot}:

\begin{itemize}[leftmargin=3mm]
    \item \textbf{Lack of Clinical Knowledge (36\%)} : The most frequent errors resulted from insufficient specialized medical knowledge, such as misidentifying anatomical structures or misunderstanding specific medical procedures.\vspace{-2mm}
    \item \textbf{Lack of Knowledge in Regulations (34\%)} : Errors often arose due to a lack of understanding of Korean healthcare laws and regulations. For example, models incorrectly assigned protocols unique to the Korean healthcare system. \vspace{-2mm}
    \item \textbf{Clinical Reasoning Errors (24\%)} : Models failed to apply clinical reasoning, even when they possessed the necessary medical information. They overlooked key clinical signs or failed to integrate all patient data, leading to incorrect diagnoses.\vspace{-3mm}
    \item \textbf{Other Errors (6\%)} : Other error types included Answer Annotation Errors (2\%) and Question Annotation Errors (4\%).
\end{itemize}

\subsection{Correlation Measurment}

    \begin{table}[t]
    \centering
    \setlength{\heavyrulewidth}{1.0pt}
    \resizebox{1.0\columnwidth}{!}{
        \newcommand{\mytable}{
\begin{tabular}{c|ccccc}
\toprule
\multicolumn{1}{c|}{\textbf{Spearman ($\rho$)}} & MedQA & IFEval & BBH & GPQA & MMLU-PRO \\ \midrule
KorMedMCQA & \textbf{0.763} & 0.769 & 0.728 & 0.535 & 0.754 \\
MedQA & - & 0.840 & 0.909 & 0.737 & 0.928 \\ \midrule
\multicolumn{1}{c|}{\textbf{Kendall ($\tau$)}} & MedQA & IFEval & BBH & GPQA & MMLU-PRO \\ \midrule
KorMedMCQA & \textbf{0.586} & 0.611 & 0.583 & 0.393 & 0.581 \\
MedQA & - & 0.683 & 0.746 & 0.549 & 0.805 \\ \bottomrule
\end{tabular}
}
            
    }
    \caption{
Correlation analysis (Spearman and Kendall) between model scores on KorMedMCQA, MedQA, and other benchmarks (IFEval, BBH, GPQA, MMLU-PRO). 
    }
    \label{correlation}
    \end{table}

In this section, we provide a deeper analysis to confirm that MedQA cannot serve as a proxy for the Korean healthcare environment.
Specifically, we measured the Spearman($\rho$) and Kendall($\tau$) correlations between model performance on KorMedMCQA and MedQA.
We then compared these results with other benchmark scores reported by the Hugging Face Open-LLM Leaderboard \citep{open-llm-leaderboard}, including IFEval \citep{zhou2023instruction}, BBH \citep{suzgun2022challenging}, GPQA \citep{rein2023gpqa}, and MMLU-PRO \citep{wang2024mmlu}.

The results, summarized in Table \ref{correlation}, reveal two key insights. 
First, the correlation between KorMedMCQA and MedQA is lower than the correlation between MedQA and most of other benchmarks in completely different domains, such as MMLU-PRO and BBH. 
This suggests that MedQA and KorMedMCQA represents two distinct medical contexts, indicating that model performance measured with MedQA is not a reliable proxy for model performance on KorMedMCQA.
Second, the correlations between KorMedMCQA and English benchmarks (IFEval, BBH, GPQA, MMLU-Pro) are consistently lower than the correlations between MedQA and those same benchmarks. 
These findings further emphasize the need for language-specific benchmarks to capture nuances and challenges unique to non-English medical tasks.

\section{Conclusion}
This study presents KorMedMCQA, a dataset of Korean medical licensing examination questions for doctors, nurses, pharmacists, and dentists. 
The dataset covers a comprehensive range of medical subjects and reflects the distinct linguistic, regulatory, and clinical landscapes of Korea’s healthcare system. 
We evaluate 59 diverse large language models using KorMedMCQA, and through correlation analysis, find that performance on this benchmark does not closely align with that on MedQA, underscoring the importance of region-specific medical benchmarks for robust assessments of real-world clinical applicability.
By making KorMedMCQA publicly available, we aim to facilitate extensive research into the development of models capable of navigating the complexities of Korean healthcare
environments.

\section{Limitation}

One limitation of this work is the potential risk of data contamination.
Since this is the first time Korean healthcare examination data has been publicly released, there is no immediate concern about contamination in the evaluations conducted in this study.
However, as the dataset becomes publicly available, future evaluations may risk biased assessments due to prior test exposure.
To address this, we plan annual test set updates with newly released questions, ensuring unbiased evaluations.


\bibliography{custom}

\newpage

\appendix

\onecolumn

\section{Data}
\label{sec:appendix1}

\begin{figure*}[!ht]
    \centering
    \includegraphics[width=1.0\textwidth,trim={60 10 65 2}, clip]{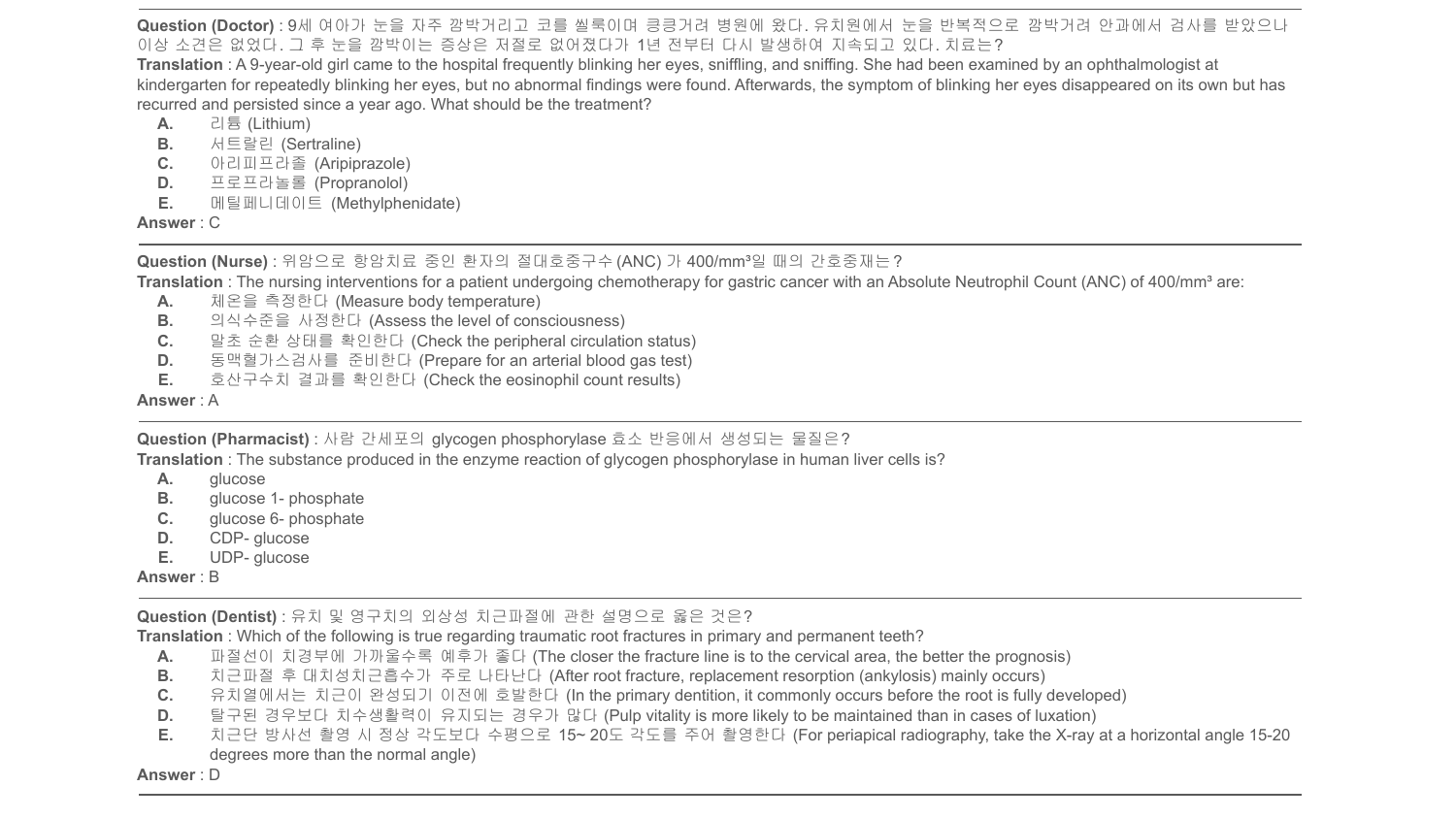}
    \caption{Sample questions of KorMedMCQA dataset respectively for doctor,
    nurse, pharmacist, and dentist license examination. Note that the translation has been added only to this figure to aid the understanding of global readers, and the actual data is not translated.}
    \label{appendix:figure1}
    \end{figure*}

This appendix provides supplementary details about the KorMedMCQA dataset. 
Figure \ref{appendix:figure1} showcases sample questions, offering insight into the dataset's structure and content.
Table \ref{appendix:split_number} details the dataset's statistics, highlighting the distribution of entries across splits (train, dev, test) and years.
Additionally, Table \ref{appendix:subjects} outlines the subjects included in the licensing examinations for doctors, nurses, pharmacists, and dentists, emphasizing the comprehensive range of topics covered by the dataset.

    \begin{table*}[!h]
    \centering
    \begin{minipage}{0.38\textwidth}
    \centering
    \setlength{\heavyrulewidth}{1.0pt}
    \resizebox{\columnwidth}{!}{
        \newcommand{\mytable}{\begin{tabular}{c|c|cccc}
\toprule
Split & Year & Doctor & Nurse & Pharm & Dentist \\ \midrule
\multirow{9}{*}{Train} & 2012 & 309 & - & - & - \\
 & 2013 & 257 & - & - & - \\
 & 2014 & 242 & - & - & - \\
 & 2015 & 226 & - & - & - \\
 & 2016 & 203 & - & - & - \\
 & 2017 & 179 & - & - & - \\
 & 2018 & 164 & - & - & - \\
 & 2019 & 163 & 291 & 318 & - \\
 & 2020 & 147 & 291 & 314 & 297 \\ \midrule
Dev & 2021 & 164 & 291 & 300 & 304 \\ \midrule
\multirow{3}{*}{Test} & 2022 & 135 & 295 & 301 & 302 \\
 & 2023 & 150 & 292 & 313 & 257 \\
 & 2024 & 150 & 291 & 271 & 252 \\ \midrule
Total & - & 2,489 & 1,751 & 1,817 & 1,412 \\ \bottomrule
\end{tabular}}
            
    }
    \caption{KorMedMCQA Statistics: Distribution of entries across splits (train, dev, test) and years for each profession (Doctor, Nurse, Pharmacist, Dentist).}
    \label{appendix:split_number}
    \end{minipage}
    \hfill
    \begin{minipage}{0.60\textwidth}
    \centering
    \setlength{\heavyrulewidth}{1.0pt}
    \resizebox{\columnwidth}{!}{
        \newcommand{\mytable}{\begin{tabular}{l|l}
\toprule
       & Subjects                                                                                                                                                                                                                                                                                                                                                             \\ \midrule
Doctor & Healthcare Laws and Regulations, General Medicine, Specialized Medicine                                                                                                                                                                                                                                                                                                                                                                                         \\ \midrule
Nurse  & \begin{tabular}[c]{@{}l@{}}Healthcare Laws and Regulations, Adult Nursing Maternal, Nursing Pediatric, \\ Nursing Community Health Nursing, Psychiatric Nursing, Nursing Management, \\ Fundamental Nursing\end{tabular}                                                                                                                                                                                                                                        \\ \midrule
Pharm  & \begin{tabular}[c]{@{}l@{}}Biochemistry, Molecular Biology, Microbiology, Immunology, Pharmacology, \\ Preventive Medicine, Pathophysiology, Physical Pharmacy, Synthetic Chemistry, \\ Medicinal Chemistry, Pharmaceutical Analysis, Pharmaceutics, Pharmacognosy, \\ Herbal Medicine, Pharmacotherapy, Pharmacy Practice, Drug Manufacturing, \\ Drug Quality Control, Pharmacy Administration and Management (Social Pharmacy), \\ Pharmacy Law \end{tabular} \\  \midrule

Dentist  & \begin{tabular}[c]{@{}l@{}}Oral Medicine, Prosthodontics, Pediatric Dentistry, Orthodontics, Oral Pathology, \\ Oral Biology, Oral and Maxillofacial Radiology, Periodontology, Oral and Maxillofacial Surgery, \\ Conservative Dentistry, Dental Public Health, Pharmaceutics, Pharmacognosy, \\ Herbal Medicine, Dental Materials, Health and Medical Law Regulations  \end{tabular} \\

\bottomrule 
\end{tabular}}
            
    }
    \caption{Subjects Covered in KorMedMCQA: List of topics included for each professional subset—Doctor, Nurse, Pharmacist, and Dentist.}
    \label{appendix:subjects}
    \end{minipage}
    \end{table*}

\newpage

\section{Evaluation} \label{appendix:evaluation}

\begin{figure*}[h]
    \centering
    \includegraphics[width=0.8\textwidth,trim={155 10 210 25}, clip]{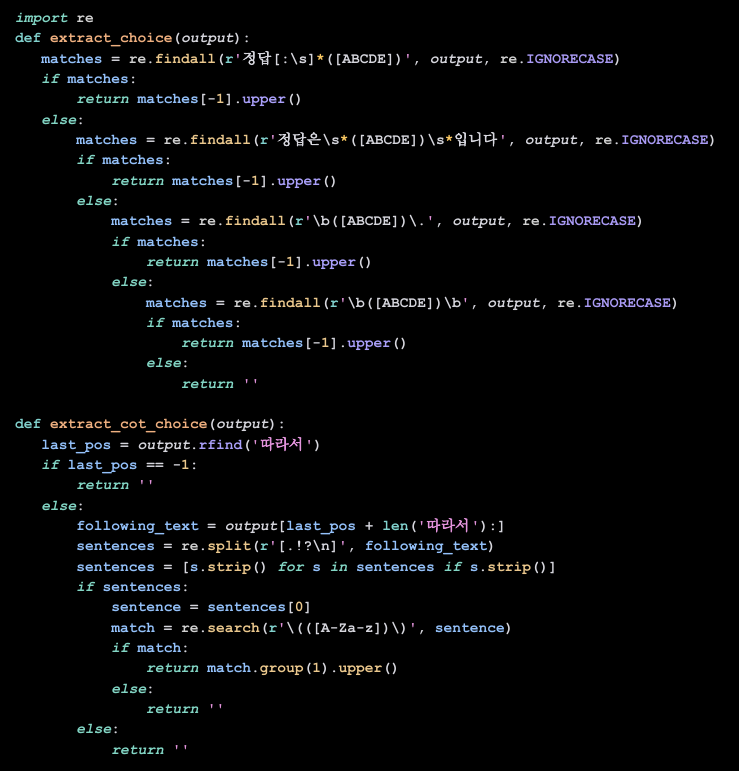}
    \caption{Regular expressions used to extract answers from model outputs when evaluating KorMedMCQA.}
    \vspace{-3mm}
    \label{appendix:reg}
    \end{figure*}

For the experiments, proprietary models were accessed via their respective APIs, while open-source models were evaluated using the LM-Harness framework \citep{eval-harness}.
Open-source models were evaluated on 8 NVIDIA A100 GPUs, each with 48 GB of memory, with the random seed set to 0.
All models, both proprietary and open-source, were tested with a temperature setting of 0 and employed a 5-shot prompting approach.
Figure \ref{appendix:reg} illustrates the regular expressions used to extract answers from the model outputs.
The function \textit{extract\_choice} was used to extract the answer index in direct answer generation, while the function \textit{extract\_cot\_choice} was employed for extracting the answer index in the CoT (Chain of Thought) setting.
Tables \ref{appendix/table3} to \ref{appendix/table10} detail the exact prompts used for the CoT evaluation.
Note that the CoT examples were crafted and validated by professionals in their respective domains: doctors, nurses, pharmacists, and dentists.
The prompts used for direct answer generation were identical to those in the CoT evaluation but excluded the reasoning process.
\begin{itemize}
    \item Table \ref{appendix/table3}: KormedMCQA doctor CoT prompt, and Table \ref{appendix/table4}: its translated version.
    \item Table \ref{appendix/table5}: KormedMCQA nurse CoT prompt, and Table \ref{appendix/table6}: its translated version.
    \item Table \ref{appendix/table7}: KormedMCQA pharmacist CoT prompt, and Table \ref{appendix/table8}: its translated version.
    \item Table \ref{appendix/table9}: KormedMCQA dentist CoT prompt, and Table \ref{appendix/table10}: its translated version.
\end{itemize}

\newpage

    \begin{table}[!h]
    \centering
    \setlength{\heavyrulewidth}{0.8pt}
    
    \newcolumntype{l}{>{\raggedright\arraybackslash}p{\columnwidth}}
    \small

    \newcommand{\mytable}{\begin{tabular}{l}
\toprule
\textbf{질문:} 13개월 남아가, 형이 하루 전 홍역으로 진단을 받아서 병원에 왔다. 평소 건강하며 현재 특별한 증상은 없다. 맥박 100회/분, 호흡 28회/분, 체온 36.7℃이다. 예방접종 기록은 다음과 같다. 조치는? BCG 1회B형간염백신 3회 DTaP-폴리오-Hib 혼합백신 3회 폐렴구균단백결합백신 4회 일본뇌염백신 1회 \\
\textbf{보기:} A. 경과관찰 B. 비타민A C. 단클론항체 D. 면역글로불린 E. 홍역- 볼거리- 풍진 백신 \\
\textbf{정답:} 차근차근 생각해봅시다. 13개월 남아는 아직 면역체계가 완전히 발달하지 않았으며, 형이 하루 전에 홍역으로 진단을 받았기 때문에 홍역 노출 가능성이 있습니다. 홍역은 전염성이 매우 강하며, 예방접종을 통해 예방이 가능합니다. 예방접종 기록을 살펴보면, BCG, B형간염백신, DTaP-폴리오-Hib 혼합백신, 폐렴구균단백결합백신, 일본뇌염백신이 접종되었으나 홍역-볼거리-풍진(MMR) 백신은 아직 접종되지 않았습니다. 그에 따라,이 아이는 아직 홍역에 대한 면역이 형성되지 않았을 가능성이 높습니다. 홍역 백신 미접종자에게 홍역 노출 후 빠르게 면역을 형성하기 위해 72시간 내 MMR 백신을 접종하는 것이 적절한 조치입니다. 따라서, 정답은 (E) 입니다. \\
\break
\textbf{질문:} 55세 여자가 2일 전부터 열이 난다며 병원에 왔다. 며칠 전부터 평소보다 소변을 자주 본다고 한다. 1개월 전에도 비슷한 증상이 있었다고 한다. 혈압 130/80 mmHg, 맥박 105회/분, 호흡 18회/분, 체온 38.5℃이다. 왼쪽 갈비척추각에 압통이 있다. 소변검사 결과는 다음과 같다. 가능성이 높은 원인미생물은? 백혈구 10～20/고배율시야, 적혈구 0～5/고배율시야 \\
\textbf{보기:} A. 대장균(Escherichia coli ) B. 창자알균(Enterococcus faecalis) C. 프로테우스미라빌리스(Proteus mirabilis) D. 우레아플라스마우레알리티쿰(Ureaplasma urealyticum ) E. 스타필로코쿠스사프로피티쿠스(Staphylococcus saprophyticus) \\
\textbf{정답:} 차근차근 생각해봅시다. 55세 여자는 2일 전부터 열이 나고, 소변을 자주 보는 증상이 있습니다. 또한 왼쪽 갈비척추각에 압통이 있으며, 체온이 38.5℃로 발열이 동반되고 있습니다. 이 환자는 소변검사에서 백혈구가 많이 검출되어 요로감염의 가능성이 높습니다. 특히, 1개월 전에도 비슷한 증상이 있었다고 하므로, 반복적인 요로감염의 가능성도 고려해야 합니다. 요로감염의 주요 원인균은 대장균(Escherichia coli)입니다. 대장균은 요로감염의 가장 흔한 원인균으로, 특히 여성에서 빈번하게 발생합니다. 환자의 소변검사 결과와 임상 증상을 종합해 볼 때, 가장 가능성이 높은 원인 미생물은 대장균입니다. 따라서, 정답은 (A) 입니다. \\
\break
\textbf{질문:} 8세 남아가 쉽게 짜증을 내고 친구를 자주 때려 병원에 왔다. 학교에서는 선생님의 지시를 따르지 못하고 집에서 자주 울고 떼를 썼으며 밤늦게까지 안 자서 엄마가 감당하기 힘들었다. 5세 때 한 차례 경련을 했고 이후로 괜찮아 치료를 받지 않았다. 지능지수는 49이다. 치료는? \\
\textbf{보기:} A. 졸피뎀 B. 도네페질 C. 리스페리돈 D. 아고멜라틴 E. 알프라졸람 \\
\textbf{정답:} 차근차근 생각해봅시다. 8세 남아는 쉽게 짜증을 내고 공격적인 행동을 보이며, 학교와 가정에서 문제를 일으키고 있습니다. 또한, 밤늦게까지 잠을 자지 못해 엄마가 감당하기 어려운 상태입니다. 이 아이는 현재 지능지수는 49로 지적장애가 있는 것으로 보입니다. 이러한 행동 문제는 지적장애와 관련된 행동 장애의 일부일 수 있으며, 치료가 필요합니다. 지적장애와 관련된 행동 장애의 치료는 행동치료, 항정신병 약물 등이 있고 또한 ADHD와의 감별이 필요합니다. 이중 보기에 있는 리스페리돈은 항정신병 약물로, 특히 공격적인 행동이나 자해 등의 문제 행동을 줄이는 데 사용될 수 있습니다. 이 약물은 지적장애가 있는 아이들의 행동 문제를 완화시키는 데 효과적입니다. 따라서, 이 경우 적절한 치료는 리스페리돈이 될 것입니다. 따라서, 정답은 (C) 입니다. \\
\break
\textbf{질문:} 50세 남자가 10일 전부터 서서히 배가 불러온다고 병원에 왔다. 누나와 형이 간질환으로 약을 복용 중이라고 한다. 혈압 110/70 mmHg, 맥박 80회/분, 호흡 18회/분, 체온 36.7℃이다. 결막은 창백하고 공막에 황달이 있다. 앞가슴에 거미혈관종이 보인다. 복부는 팽만하고 이동둔탁음이 있다. 복부에 압통이나 반동압통은 없다. 양쪽 정강뼈앞 오목부종이 있다. 검사 결과는 다음과 같다. 복부 컴퓨터단층촬영 사진이다. 치료는? 혈액: 백혈구 4,700/mm\^3, 혈색소 10.4 g/dL, 혈소판 59,000/mm\^3 프로트롬빈시간(INR) 2.1, 알부민 1.8 g/dL,총빌리루빈 4.2 mg/dL, 아스파르테이트아미노전달효소 69 U/L 알라닌아미노전달효소 18 U/L HBsAg (+), anti-HBs (-), anti-HCV (-) 알파태아단백질 3.9 ng/mL (참고치, <8.5) 복수: 총단백질 1.2 g/dL, 알부민 0.4 g/dL,백혈구 100/mm\^3, 아데노신탈아미노효소(ADA) 9 U/L \\
\textbf{보기:} A. 리팜핀 B. 세포탁심 C. 락툴로오스 D. 스피로놀락톤 E. 메트로니다졸 \\
\textbf{정답:} 차근차근 생각해봅시다. 50세 남자는 10일 전부터 배가 불러오는 증상으로 병원에 왔습니다. 환자의 가족력에서 간질환이 있으며, 현재 환자에게 결막 창백, 황달, 거미혈관종, 복부 팽만, 이동둔탁음, 그리고 오목부종과 같은 간경변의 징후들이 보입니다. 검사 결과에서 저알부민혈증, 고빌리루빈혈증, 그리고 혈소판 감소증이 나타나고 있으며, HBsAg 양성으로 B형 간염과 관련된 만성 간질환이 있을 가능성이 높습니다. 복수 검사 결과, 저단백질 복수와 저알부민 농도가 확인되었습니다. 또한, 복수의 백혈구 수치가 낮아 감염의 가능성이 낮으며, ADA 수치도 결핵성 복수를 시사하지 않습니다. 이러한 소견은 간경변으로 인한 복수의 발생을 의미하며, 이로 인해 복수 관리가 필요합니다. 이 상황에서는 이뇨제인 스피로놀락톤이 복수를 조절하는 데 가장 효과적인 치료법이 될 수 있습니다. 따라서, 정답은 (D) 입니다. \\
\break
\textbf{질문:} 35세 여자가 2주 전부터 배가 불러온다며 병원에 왔다. 2개월 동안 체중이 4 kg 줄었다고 한다. 혈압 110/70 mmHg, 맥박 97회/분, 호흡 21회/분, 체온 38.1℃이다. 의식은 명료하다. 배는 팽만하고 이동둔탁음이 있다. 배에 가벼운 압통은 있으나 반동압통은 없다. 검사 결과는 다음과 같다. 원인은?혈액: 혈색소 11.2g/dL, 백혈구 4,200/mm\^3, 혈소판 169,000/mm\^3 총단백질 6.0 g/dL, 알부민 3.3 g/dL, 총빌리루빈 1.2 mg/dL, 아스파르테이트아미노전달효소 34 U/L 알라닌아미노전달효소 21 U/L 아밀라아제 61 U/L, 리파제 30 U/L 복수: 백혈구 1,500/mm\^3 (중성구 5\%, 림프구 90\%) 단백질 3.1 g/dL, 알부민 2.5 g/dL,아데노신탈아미노효소(ADA) 85 U/L 소변: 단백질 (-), 적혈구 0～2/고배율시야 \\
\textbf{보기:} A. 간경화증 B. 복막결핵 C. 콩팥증후군 D. 급성이자염 E. 복막암종증 \\
\textbf{정답:} 차근차근 생각해봅시다. 35세 여자는 2주 전부터 배가 불러오는 증상으로 병원에 왔고, 최근 2개월 동안 체중이 4 kg 줄어들었습니다. 체온이 38.1℃로 발열이 동반되었고, 복부 팽만과 이동둔탁음이 있습니다. 검사 결과를 보면, 혈액 검사에서 총단백질과 알부민 수치는 정상 하한선에 가깝지만, 복수에서 단백질과 알부민 수치가 상대적으로 높고, ADA(아데노신탈아미노효소) 수치가 85 U/L로 상승되어 있습니다. 복수의 백혈구 수치도 증가했으며 림프구 비율이 높습니다. 이러한 소견들은 결핵에 의한 복막염, 즉 복막결핵을 강력히 시사합니다. ADA 수치의 상승은 특히 결핵성 복막염에서 흔히 관찰됩니다. 따라서, 정답은 (B) 입니다. \\
\bottomrule
\end{tabular}}

    \caption{
KormedMCQA Doctor CoT Prompt (Original Korean Version)
    }
    \label{appendix/table3}
    \end{table}

\newpage

    \begin{table}[!h]
    \centering
    \setlength{\heavyrulewidth}{0.8pt}
    
    \newcolumntype{l}{>{\raggedright\arraybackslash}p{\columnwidth}}
    \small

    \newcommand{\mytable}{\begin{tabular}{l}
\toprule
\textbf{Question:} A 13-month-old boy visits the hospital after his older brother was diagnosed with measles the day before. The boy is usually healthy and currently has no specific symptoms. His pulse is 100 beats/min, respiratory rate is 28 breaths/min, and temperature is 36.7℃. His vaccination record is as follows. What is the appropriate action? BCG: 1 dose, Hepatitis B vaccine: 3 doses, DTaP-IPV-Hib combined vaccine: 3 doses, Pneumococcal conjugate vaccine: 4 doses, Japanese encephalitis vaccine: 1 dose. \\
\textbf{Choices:} A. Observation B. Vitamin A C. Monoclonal antibody D. Immunoglobulin E. Measles-Mumps-Rubella vaccine \\
\textbf{Answer:} Let’s think step by step. The 13-month-old boy's immune system is not yet fully developed. Considering that his older brother was diagnosed with measles the day before, there is a potential exposure to measles. Measles is highly contagious and preventable through vaccination. Reviewing the vaccination record, the child has received BCG, Hepatitis B vaccine, DTaP-IPV-Hib combined vaccine, Pneumococcal conjugate vaccine, and Japanese encephalitis vaccine, but the Measles-Mumps-Rubella (MMR) vaccine has not been administered yet. Therefore, it is highly likely that the child does not have immunity to measles. Administering the MMR vaccine within 72 hours of measles exposure is an appropriate measure to rapidly induce immunity. Thus, the correct answer is (E). \\
\break
\textbf{Question:} A 55-year-old woman visits the hospital complaining of fever for 2 days. She also reports frequent urination over the past few days and mentions a similar episode one month ago. Her blood pressure is 130/80 mmHg, pulse is 105 beats/min, respiratory rate is 18 breaths/min, and temperature is 38.5℃. She has tenderness in the left costovertebral angle. A urinalysis shows the following. What is the most likely causative microorganism? White blood cells: 10-20/high power field, Red blood cells: 0-5/high power field. \\
\textbf{Choices:} A. Escherichia coli B. Enterococcus faecalis C. Proteus mirabilis D. Ureaplasma urealyticum E. Staphylococcus saprophyticus \\
\textbf{Answer:} Let’s think step by step. The 55-year-old woman has fever and frequent urination, along with tenderness in the left costovertebral angle and a temperature of 38.5℃. The urinalysis shows an increased white blood cell count, suggesting a urinary tract infection (UTI). Given the history of a similar episode one month ago, recurrent UTI should also be considered. The most common causative microorganism of UTI is Escherichia coli. E. coli is the leading cause of UTI, especially in women. Based on the urinalysis and clinical symptoms, the most likely causative microorganism is E. coli. Thus, the correct answer is (A). \\
\break
\textbf{Question:} An 8-year-old boy is brought to the hospital for frequent irritability and hitting friends. At school, he fails to follow the teacher's instructions, cries frequently at home, throws tantrums, and stays awake late at night, overwhelming his mother. He had a seizure at age 5 but has not received treatment since then. His IQ is 49. What is the treatment? \\
\textbf{Choices:} A. Zolpidem B. Donepezil C. Risperidone D. Agomelatine E. Alprazolam \\
\textbf{Answer:} Let’s think step by step. The 8-year-old boy exhibits irritability, aggressive behavior, and behavioral problems both at school and home. Additionally, he has difficulty sleeping, causing significant stress for his mother. His IQ of 49 suggests an intellectual disability. Behavioral issues in children with intellectual disabilities can include aggression and self-injurious behavior, which may require treatment. Among the choices, Risperidone, an antipsychotic medication, is effective in reducing aggression and severe behavioral problems in children with intellectual disabilities. Therefore, the appropriate treatment in this case is Risperidone. Thus, the correct answer is (C). \\
\break
\textbf{Question:} A 50-year-old man visits the hospital complaining of gradual abdominal swelling over the past 10 days. His siblings are on medication for liver disease. His blood pressure is 110/70 mmHg, pulse is 80 beats/min, respiratory rate is 18 breaths/min, and temperature is 36.7℃. He has pale conjunctiva, jaundice in the sclera, and spider angiomas on his chest. His abdomen is distended with shifting dullness but no tenderness or rebound tenderness. He has pitting edema in both shins. Laboratory and imaging results are as follows. What is the treatment? Blood: WBC 4,700/mm\^3, Hemoglobin 10.4 g/dL, Platelet count 59,000/mm\^3, Prothrombin time (INR) 2.1, Albumin 1.8 g/dL, Total bilirubin 4.2 mg/dL, AST 69 U/L, ALT 18 U/L, HBsAg (+), anti-HBs (-), anti-HCV (-), Alpha-fetoprotein 3.9 ng/mL (reference, <8.5). Ascitic fluid: Total protein 1.2 g/dL, Albumin 0.4 g/dL, WBC 100/mm\^3, Adenosine deaminase (ADA) 9 U/L. \\
\textbf{Choices:} A. Rifampin B. Cefotaxime C. Lactulose D. Spironolactone E. Metronidazole \\
\textbf{Answer:} Let’s think step by step. The 50-year-old man presents with abdominal distension, family history of liver disease, and signs of liver cirrhosis such as jaundice, spider angiomas, and ascites. Laboratory results indicate hypoalbuminemia, hyperbilirubinemia, and thrombocytopenia. HBsAg positivity suggests chronic liver disease related to hepatitis B. Ascitic fluid analysis shows low protein and albumin levels without significant infection or tuberculosis indicators. These findings suggest cirrhosis-induced ascites. Spironolactone, a diuretic, is effective in managing ascites in this context. Thus, the correct answer is (D). \\
\break
\textbf{Question:} A 35-year-old woman visits the hospital for abdominal swelling lasting 2 weeks. She has lost 4 kg over the past 2 months. Her blood pressure is 110/70 mmHg, pulse is 97 beats/min, respiratory rate is 21 breaths/min, and temperature is 38.1℃. She is alert. Her abdomen is distended with shifting dullness and mild tenderness but no rebound tenderness. Laboratory results are as follows. What is the cause? Blood: Hemoglobin 11.2 g/dL, WBC 4,200/mm\^3, Platelet count 169,000/mm\^3, Total protein 6.0 g/dL, Albumin 3.3 g/dL, Total bilirubin 1.2 mg/dL, AST 34 U/L, ALT 21 U/L, Amylase 61 U/L, Lipase 30 U/L. Ascitic fluid: WBC 1,500/mm\^3 (5\% neutrophils, 90\% lymphocytes), Protein 3.1 g/dL, Albumin 2.5 g/dL, ADA 85 U/L. Urine: Protein (-), RBC 0–2/high power field. \\
\textbf{Choices:} A. Liver cirrhosis B. Tuberculous peritonitis C. Nephrotic syndrome D. Acute pancreatitis E. Peritoneal carcinomatosis \\
\textbf{Answer:} Let’s think step by step. The 35-year-old woman presents with abdominal distension, fever, and significant weight loss. Her ascitic fluid analysis reveals high protein and albumin levels and elevated ADA (85 U/L), with a predominance of lymphocytes. These findings are strongly suggestive of tuberculous peritonitis. Elevated ADA is a hallmark of tuberculosis-related peritonitis. Thus, the correct answer is (B). \\
\bottomrule
\end{tabular}
}

    \caption{
KormedMCQA Doctor CoT Prompt (Translated English Version)
    }
    \label{appendix/table4}
    \end{table}

\newpage

    \begin{table}[!h]
    \centering
    \setlength{\heavyrulewidth}{0.8pt}
    
    \newcolumntype{l}{>{\raggedright\arraybackslash}p{\columnwidth}}
    \small

    \newcommand{\mytable}{\begin{tabular}{l}
\toprule
\textbf{질문:} 옥시토신 유도분만 중인 40주 초산부가 통증을 호소하여 사정한 결과는 다음과 같다. 우선적인 간호중재는?• 자궁경관: 개대 2 cm, 소실 70\% • 자궁수축: 간격 2분 미만, 기간 90초 이상, 강도 100 mmHg \\
\textbf{보기:} A. 관장 시행 B. 양막절개술 실시 C. 자궁근이완제 투여 D. 옥시토신 투여 중단 E. 프로스타글란딘 질정 투여 \\
\textbf{정답:} 차근차근 생각해봅시다. 40주 초산부가 옥시토신 유도분만 중에 있습니다. 현재 자궁경관의 개대는 2 cm, 소실은 70\%이며, 자궁수축 간격은 2분 미만이고, 수축 기간은 90초 이상입니다. 강도는 100 mmHg로 매우 강한 상태입니다. 이 경우 자궁수축이 과도하여 태아와 산모에게 위험이 될 수 있습니다. 옥시토신은 자궁수축을 유도하기 위해 사용되지만, 과도한 수축은 자궁의 혈류를 감소시켜 태아의 산소 공급에 영향을 줄 수 있습니다. 현재 이 산모의 자궁수축이 과도하므로, 옥시토신 투여를 중단하거나 감량하는 것이 필요합니다. 따라서, 정답은 (D) 입니다. \\
\break
\textbf{질문:} 심부전 환자의 사정결과가 다음과 같을 때 우선적인 중재는?• 허약, 심한 피로, 야뇨 • 거품이 있는 분홍빛의 객담, 기좌호흡 • 맥박 110회/분, 호흡 30회/분 \\
\textbf{보기:} A. 수분섭취 권장 B. 산소포화도 측정 C. 객담배양검사 실시 D. 트렌델렌부르크자세 유지 E. 복도 걷기와 같은 활동 격려 \\
\textbf{정답:} 차근차근 생각해봅시다. 심부전 환자의 사정 결과를 보면, 허약과 심한 피로, 야뇨가 있으며, 거품이 있는 분홍빛의 객담과 기좌호흡이 나타나고 있습니다. 또한, 맥박은 110회/분, 호흡은 30회/분으로 빈맥과 빈호흡이 관찰됩니다. 이러한 증상들은 폐에 액체가 축적되어 발생하는 폐부종을 시사합니다. 폐부종은 산소 공급에 심각한 영향을 미칠 수 있으며, 즉각적인 중재가 필요합니다. 이 환자의 경우 산소 공급이 필요할 수 있으므로, 우선 산소포화도를 측정하여 산소 요구량을 평가하는 것이 중요합니다. 따라서, 정답은 (B) 입니다. \\
\break
\textbf{질문:} 호흡곤란으로 입원한 환자의 사정결과가 다음과 같을 때 우선적인 중재는?• 3개월 전 승모판대치술을 받음• 목정맥 팽창, 체중 증가• 심박출률(EF) 45\%, 산소포화도 86\% • 혈압 90/60 mmHg, 맥박 102회/분, 호흡 26회/분, 체온 37.5℃ \\
\textbf{보기:} A. 산소 투여 B. 배변완화제 투여 C. 중등도 활동 격려 D. 혈액배양검사 실시 E. 비위관 영양 공급 \\
\textbf{정답:} 차근차근 생각해봅시다. 환자는 3개월 전 승모판대치술을 받은 상태이며, 현재 호흡곤란으로 입원 중입니다. 사정결과를 보면 목정맥 팽창과 체중 증가가 나타나고 있으며, 심박출률(EF)이 45\%로 감소한 상태입니다. 또한, 산소포화도가 86\%로 저하되어 있고, 혈압은 90/60 mmHg로 낮으며, 맥박과 호흡도 증가한 상태입니다. 이 환자의 상황은 심부전의 증상으로 볼 수 있으며, 특히 산소포화도가 86\%로 낮은 상태는 산소 공급이 부족함을 의미합니다. 즉각적인 산소 공급이 필요할 수 있으므로, 우선 산소를 투여하는 것이 중요합니다. 따라서, 정답은 (A) 입니다. \\
\break
\textbf{질문:} 외래 진료를 받고 있는 환자의 사정결과가 다음과 같을 때 간호중재는? • 전신 소양증, 만성 두통• 적혈구 800만/mm\^3, 헤모글로빈 20 g/dL,헤마토크리트 60\%, 혈소판 45만/mm\^3 \\
\textbf{보기:} A. 침상안정을 한다. B. 철분제제를 투여한다. C. 치실을 사용하여 치아를 관리한다. D. 앉아있을 때 다리를 내리도록 한다. E. 수분을 하루 3 L 이상 섭취하도록 격려한다. \\
\textbf{정답:} 차근차근 생각해봅시다. 외래 진료를 받고 있는 환자의 사정 결과를 보면 전신 소양증과 만성 두통을 호소하고 있으며, 혈액 검사 결과 적혈구 수치가 800만/mm³로 매우 높고, 헤모글로빈이 20 g/dL, 헤마토크리트가 60\%로 증가되어 있습니다. 혈소판 수치도 45만/mm³로 높게 나타났습니다. 이러한 결과는 적혈구 증가증(polycythemia)를 시사하며, 이는 혈액이 끈적해지고 혈전 발생 위험이 증가할 수 있는 상태입니다. 혈액의 점도를 낮추고 혈액 순환을 원활하게 하기 위해서는 충분한 수분 섭취가 필요합니다. 따라서, 이 환자에게는 수분을 하루 3L 이상 섭취하도록 격려하는 것이 중요합니다. 따라서, 정답은 (E) 입니다. \\
\break
\textbf{질문:} 패혈증 환자의 사정결과가 다음과 같을 때 우선적인 중재는?• 심한 호흡곤란, 의식수준 저하, 창백한 피부• 맥박 110회/분, 수축기 혈압 70 mmHg • 흉부 X선 검사 시 양쪽 폐 부위가 하얗게 나타남• 동맥혈가스분석: 이산화탄소분압 70 mmHg 산소분압 50 mmHg, pH 7.18산소포화도 65\% \\
\textbf{보기:} A. 비위관 삽입 B. 항응고제 투여 C. 인공호흡기 적용 D. 폐기능검사 실시 E. 흉부물리요법 제공 \\
\textbf{정답:} 차근차근 생각해봅시다. 이 환자는 패혈증으로 인해 심한 호흡곤란과 의식수준 저하, 창백한 피부를 보이고 있습니다. 맥박은 110회/분으로 빈맥이 나타나고 있으며, 수축기 혈압이 70 mmHg로 매우 낮은 상태입니다. 흉부 X선 검사에서는 양쪽 폐 부위가 하얗게 나타나, 폐에 심각한 이상이 있음을 시사합니다. 또한, 동맥혈가스분석 결과 이산화탄소분압이 70 mmHg로 상승하고, 산소분압이 50 mmHg로 저하되어 있으며, pH는 7.18로 산증 상태입니다. 산소포화도는 65\%로 매우 낮아, 환자는 심각한 저산소증 상태에 있습니다. 이 상황에서는 즉각적인 산소 공급이 필수적이며, 환자가 자발적으로 호흡을 유지하기 어려운 상태이므로 인공호흡기를 적용하여 산소를 공급하고 환자의 호흡을 보조하는 것이 필요합니다. 따라서, 정답은 (C) 입니다. \\
\bottomrule
\end{tabular}
}

    \caption{
KormedMCQA Nurse CoT Prompt (Original Korean Version)
    }
    \label{appendix/table5}
    \end{table}

\newpage

    \begin{table}[!h]
    \centering
    \setlength{\heavyrulewidth}{0.8pt}
    
    \newcolumntype{l}{>{\raggedright\arraybackslash}p{\columnwidth}}
    \small

    \newcommand{\mytable}{\begin{tabular}{l}
\toprule
\textbf{Question:} A 40-week primipara undergoing oxytocin induction of labor complains of pain, and the following assessment results are obtained. What is the priority nursing intervention? • Cervical dilation: 2 cm, effacement 70\% • Uterine contractions: intervals less than 2 minutes, duration over 90 seconds, intensity 100 mmHg \\
\textbf{Options:} A. Perform an enema B. Perform amniotomy C. Administer uterine relaxant D. Discontinue oxytocin infusion E. Administer prostaglandin suppository \\
\textbf{Answer:} Let’s think step by step. The patient is a 40-week primipara undergoing oxytocin-induced labor. The cervical dilation is 2 cm with 70\% effacement. Uterine contractions occur at intervals of less than 2 minutes, last over 90 seconds, and have an intensity of 100 mmHg, indicating excessive contractions. Oxytocin is used to induce uterine contractions; however, excessive contractions can reduce uterine blood flow and affect fetal oxygenation. Given this patient's excessive uterine contractions, discontinuing or reducing the oxytocin infusion is necessary. Thus, the correct answer is (D). \\
\break
\textbf{Question:} What is the priority intervention for a heart failure patient with the following assessment findings? • Weakness, severe fatigue, nocturia • Frothy, pink-tinged sputum, orthopnea • Pulse 110 bpm, respiration 30 breaths/min \\
\textbf{Options:} A. Encourage fluid intake B. Measure oxygen saturation C. Perform sputum culture D. Maintain Trendelenburg position E. Encourage activities like hallway walking \\
\textbf{Answer:} Let’s think step by step. The heart failure patient presents with weakness, severe fatigue, and nocturia, as well as frothy, pink-tinged sputum and orthopnea. The patient exhibits tachycardia (pulse 110 bpm) and tachypnea (respiration 30 breaths/min). These findings indicate pulmonary edema, which can significantly impair oxygenation and requires immediate intervention. Measuring oxygen saturation is essential to assess oxygen needs. Thus, the correct answer is (B). \\
\break
\textbf{Question:} What is the priority intervention for a patient hospitalized with dyspnea and the following assessment findings? • Mitral valve replacement 3 months ago • Jugular vein distention, weight gain • Ejection fraction (EF) 45\%, oxygen saturation 86\% • Blood pressure 90/60 mmHg, pulse 102 bpm, respiration 26 breaths/min, temperature 37.5°C \\
\textbf{Options:} A. Administer oxygen B. Administer laxatives C. Encourage moderate activity D. Perform blood culture E. Provide nasogastric nutrition \\
\textbf{Answer:} Let’s think step by step. The patient had a mitral valve replacement 3 months ago and is hospitalized for dyspnea. Findings include jugular vein distention, weight gain, and an EF of 45\%, indicating reduced cardiac output. Oxygen saturation is 86\%, and blood pressure is 90/60 mmHg, suggesting oxygen deficiency. Immediate oxygen administration is necessary to address hypoxemia. Thus, the correct answer is (A). \\
\break
\textbf{Question:} What is the nursing intervention for an outpatient with the following assessment findings? • Generalized itching, chronic headache • RBC 8 million/mm\^3, hemoglobin 20 g/dL, hematocrit 60\%, platelets 450,000/mm\^3 \\
\textbf{Options:} A. Maintain bed rest B. Administer iron supplements C. Use dental floss for oral care D. Keep legs down while sitting E. Encourage daily fluid intake of over 3 L \\
\textbf{Answer:} Let’s think step by step. The outpatient presents with generalized itching and chronic headache. Laboratory findings show an RBC count of 8 million/mm\^3, hemoglobin of 20 g/dL, hematocrit of 60\%, and platelets of 450,000/mm\^3, indicating polycythemia. This condition increases blood viscosity and thrombotic risk. Adequate fluid intake helps reduce blood viscosity and improve circulation. Thus, the correct answer is (E). \\
\break
\textbf{Question:} What is the priority intervention for a sepsis patient with the following assessment findings? • Severe dyspnea, decreased consciousness, pale skin • Pulse 110 bpm, systolic BP 70 mmHg • Chest X-ray shows bilateral lung opacities • ABG: pCO\textsubscript{2} 70 mmHg, pO\textsubscript{2} 50 mmHg, pH 7.18, oxygen saturation 65\% \\
\textbf{Options:} A. Insert nasogastric tube B. Administer anticoagulants C. Apply mechanical ventilation D. Perform pulmonary function test E. Provide chest physiotherapy \\
\textbf{Answer:} Let’s think step by step. The sepsis patient exhibits severe dyspnea, reduced consciousness, pale skin, and systolic BP of 70 mmHg. Chest X-ray findings indicate significant lung abnormalities. ABG results show hypercapnia (pCO\textsubscript{2} 70 mmHg), hypoxemia (pO\textsubscript{2} 50 mmHg), and acidosis (pH 7.18), with oxygen saturation at 65\%. Immediate oxygenation through mechanical ventilation is required to support the patient’s compromised respiratory function. Thus, the correct answer is (C). \\
\bottomrule
\end{tabular}
}

    \caption{
KormedMCQA Nurse CoT Prompt (Translated English Version)
    }
    \label{appendix/table6}
    \end{table}
\newpage

    \begin{table}[!h]
    \centering
    \setlength{\heavyrulewidth}{0.8pt}
    
    \newcolumntype{l}{>{\raggedright\arraybackslash}p{\columnwidth}}
    \small

    \newcommand{\mytable}{\begin{tabular}{l}
\toprule
\textbf{질문:} Michaelis-Menten 속도론을 따르는 단일 기질의 효소반응에서 경쟁적 억제제를 첨가할 경우 측정된 겉보기 값이 증가하는 것은? \\
\textbf{보기:} A. equilibrium constant (K) B. initial velocity (V) C. maximum velocity (V max) D. Michaelis constant (K m) E. turnover number (k cat) \\
\textbf{정답:} 차근차근 생각해봅시다. Michaelis-Menten 속도론에서 경쟁적 억제제는 기질과 경쟁하여 효소의 활성 부위에 결합하려고 합니다. 경쟁적 억제제가 존재하면 기질이 효소에 결합하는 것이 어려워져서 반응이 진행되는 속도에 영향을 미치게 됩니다. 이 상황에서 Michaelis-Menten 상수를 살펴보면, 경쟁적 억제제가 존재할 때 Michaelis 상수(K m)가 증가하게 됩니다. 이는 효소가 기질과 결합하여 반응을 촉진시키기 위해 더 높은 농도의 기질이 필요함을 의미합니다. 하지만 최대 속도 V max는 변화하지 않습니다. 이는 기질의 농도가 매우 높아지면 억제제의 영향을 극복할 수 있기 때문입니다. 경쟁적 억제제를 첨가할 경우 측정된 겉보기 값이 증가하는 것은 Michaelis constant (K m)입니다. 따라서, 정답은 (D) 입니다. \\
\break
\textbf{질문:} Histone acetyltransferase에 의해 변형되는 histone 단백질의 아미노산 잔기는? \\
\textbf{보기:} A. alanine B. leucine C. lysine D. serine E. threonine \\
\textbf{정답:} 차근차근 생각해봅시다. Histone acetyltransferase (HAT)는 히스톤 단백질의 특정 아미노산 잔기에 아세틸기를 첨가하여 히스톤의 구조를 변화시키고, 이로 인해 DNA와 히스톤의 결합이 느슨해집니다. 이 과정은 주로 유전자 발현을 촉진하는 데 중요한 역할을 합니다. HAT가 변형시키는 주요 아미노산 잔기는 양전하를 띠고 있는 리신(lysine)입니다. 리신 잔기의 아세틸화는 히스톤의 양전하를 중화시켜, DNA가 히스톤으로부터 더 쉽게 풀리도록 돕습니다. 따라서, 정답은 (C) 입니다. \\
\break
\textbf{질문:} Phosphocholine이나 phosphoethanolamine을 포함하고 있는 지질은? \\
\textbf{보기:} A. ceramide B. cerebroside C. ganglioside GM1 D. globoside E. sphingomyelin \\
\textbf{정답:} 차근차근 생각해봅시다. Phosphocholine이나 phosphoethanolamine을 포함하고 있는 지질은 스핑고미엘린(sphingomyelin)입니다. 스핑고미엘린은 스핑고지질(sphingolipid)의 한 종류로, 특히 세포막의 구성 성분 중 하나로 중요한 역할을 합니다. 스핑고미엘린은 스핑고신(sphingosine) 백본에 포스포콜린(phosphocholine) 또는 포스포에탄올아민(phosphoethanolamine)이 결합된 구조를 가지고 있습니다. 따라서, 이 지질이 바로 스핑고미엘린입니다. 다른 선택지에 있는 지질들은 이러한 포스포콜린이나 포스포에탄올아민을 포함하고 있지 않습니다. 따라서, 정답은 (E) 입니다. \\
\break
\textbf{질문:} 다음에 해당하는 것은?• 시트르산 회로의 중간체• phosphofructokinase-1을 저해하여 당분해를 억제• acetyl-CoA carboxylase를 활성화하여 지방산 생합성을 촉진 \\
\textbf{보기:} A. α- ketoglutarate B. citrate C. malate D. oxaloacetate E. succinate \\
\textbf{정답:} 차근차근 생각해봅시다. 문제에서 주어진 단서를 살펴보면, 이 화합물은 시트르산 회로의 중간체이며, phosphofructokinase-1을 저해하여 당분해(glycolysis)를 억제하고, acetyl-CoA carboxylase를 활성화하여 지방산 생합성을 촉진한다고 합니다. 먼저, 시트르산 회로의 중간체를 떠올려봅시다. 시트르산 회로에서 중요한 중간체로는 α-ketoglutarate, citrate, malate, oxaloacetate, succinate 등이 있습니다. 두 번째 단서인 "phosphofructokinase-1을 저해하여 당분해를 억제"하는 기능을 생각해보겠습니다. 이 효소는 당분해 과정에서 매우 중요한 조절 지점으로, 이 효소의 저해는 당분해를 억제하여 세포 내 에너지 상태를 조절하는 데 중요한 역할을 합니다. 특히 citrate는 이 효소를 저해하는 것으로 잘 알려져 있습니다. 세 번째 단서인 "acetyl-CoA carboxylase를 활성화하여 지방산 생합성을 촉진"하는 역할도 citrate의 기능 중 하나입니다. citrate는 세포 내에서 acetyl-CoA를 공급하며, 지방산 생합성의 초기 단계인 acetyl-CoA carboxylase를 활성화하여 지방산 생합성을 촉진합니다. 따라서, 이러한 모든 단서를 종합하면, 이 화합물은 citrate입니다. 따라서, 정답은 (B) 입니다. \\
\break
\textbf{질문:} 병원감염(nosocomial infection) 을 매개하는 생물막(biofilm) 형성과 관련 있는 세균의 세포 외 다당질(exopolysaccharide) 구조는? \\
\textbf{보기:} A. 협막(capsule) B. 편모(flagella) C. 섬모(fimbriae) D. 성섬모(sex pili) E. 봉입체(inclusion body) \\
\textbf{정답:} 차근차근 생각해봅시다. 병원감염(nosocomial infection)은 병원 내에서 획득되는 감염을 말하며, 이 감염의 원인 중 하나로 생물막(biofilm)이 중요한 역할을 합니다. 생물막은 세균들이 표면에 부착하여 성장하면서 형성되는 복잡한 구조체로, 이 구조는 세균들이 외부 환경으로부터 보호받고 항생제 저항성을 증가시킬 수 있는 중요한 메커니즘을 제공합니다. 이 생물막 형성 과정에서 중요한 역할을 하는 것이 바로 세균의 세포 외 다당질(exopolysaccharide)입니다. 이 구조는 세균들이 표면에 부착하고 서로를 연결하며, 보호막을 형성하는 데 기여합니다. 주어진 선택지 중에서 이러한 세포 외 다당질과 관련된 구조를 찾아야 합니다. 협막(capsule)은 세균의 외부에 위치하는 다당질로 구성된 구조로, 세균이 표면에 부착하고 생물막을 형성하는 데 중요한 역할을 합니다. 또한, 협막은 면역 반응으로부터 세균을 보호하고, 항생제 저항성을 증가시키는 역할도 합니다. 따라서, 병원감염을 매개하는 생물막 형성과 관련 있는 세균의 세포 외 다당질 구조는 협막(capsule)입니다. 따라서, 정답은 (A) 입니다. \\
\bottomrule
\end{tabular}
}

    \caption{
KormedMCQA Pharmacist CoT Prompt (Original Korean Version)
    }
    \label{appendix/table7}
    \end{table}
\newpage

    \begin{table}[!h]
    \centering
    \setlength{\heavyrulewidth}{0.8pt}
    
    \newcolumntype{l}{>{\raggedright\arraybackslash}p{\columnwidth}}
    \small

    \newcommand{\mytable}{\begin{tabular}{l}
\toprule
\textbf{Question:} When a competitive inhibitor is added to an enzyme reaction following Michaelis-Menten kinetics with a single substrate, which measured apparent value increases? \\
\textbf{Options:} A. equilibrium constant (K) B. initial velocity (V) C. maximum velocity (V max) D. Michaelis constant (K m) E. turnover number (k cat) \\
\textbf{Answer:} Let’s think step by step. In Michaelis-Menten kinetics, a competitive inhibitor competes with the substrate to bind to the active site of the enzyme. When a competitive inhibitor is present, it becomes harder for the substrate to bind to the enzyme, affecting the reaction rate. Looking at the Michaelis constant (K m), it increases in the presence of a competitive inhibitor. This means a higher substrate concentration is required to achieve the same reaction rate. However, the maximum velocity (V max) remains unchanged because at high substrate concentrations, the effect of the inhibitor can be overcome. Therefore, the measured apparent value that increases with the addition of a competitive inhibitor is the Michaelis constant (K m). Thus, the correct answer is (D). \\
\break
\textbf{Question:} Which amino acid residue in histone proteins is modified by histone acetyltransferase? \\
\textbf{Options:} A. alanine B. leucine C. lysine D. serine E. threonine \\
\textbf{Answer:} Let’s think step by step. Histone acetyltransferase (HAT) modifies specific amino acid residues in histone proteins by adding an acetyl group, altering histone structure and loosening the interaction between histones and DNA. This process plays a crucial role in promoting gene expression. The primary amino acid residue modified by HAT is lysine, which is positively charged. Acetylation of lysine neutralizes its positive charge, facilitating the release of DNA from histones. Thus, the correct answer is (C). \\
\break
\textbf{Question:} Which lipid contains phosphocholine or phosphoethanolamine? \\
\textbf{Options:} A. ceramide B. cerebroside C. ganglioside GM1 D. globoside E. sphingomyelin \\
\textbf{Answer:} Let’s think step by step. Lipids containing phosphocholine or phosphoethanolamine are known as sphingomyelin. Sphingomyelin is a type of sphingolipid and plays a critical role as a component of cell membranes. It is composed of a sphingosine backbone with phosphocholine or phosphoethanolamine attached. Other lipids in the options do not contain phosphocholine or phosphoethanolamine. Thus, the correct answer is (E). \\
\break
\textbf{Question:} Which of the following matches the description? • Intermediate of the citric acid cycle • Inhibits phosphofructokinase-1 to suppress glycolysis • Activates acetyl-CoA carboxylase to promote fatty acid synthesis \\
\textbf{Options:} A. α- ketoglutarate B. citrate C. malate D. oxaloacetate E. succinate \\
\textbf{Answer:} Let’s think step by step. Based on the clues provided, the compound is an intermediate of the citric acid cycle, inhibits phosphofructokinase-1 to suppress glycolysis, and activates acetyl-CoA carboxylase to promote fatty acid synthesis. First, let’s identify the intermediates of the citric acid cycle, which include α-ketoglutarate, citrate, malate, oxaloacetate, and succinate. Second, the inhibition of phosphofructokinase-1 is a key function of citrate, which helps regulate energy metabolism. Finally, citrate activates acetyl-CoA carboxylase to promote fatty acid synthesis by supplying acetyl-CoA. Combining these clues, the compound is citrate. Thus, the correct answer is (B). \\
\break
\textbf{Question:} Which structure of bacterial extracellular polysaccharides is associated with biofilm formation in nosocomial infections? \\
\textbf{Options:} A. capsule B. flagella C. fimbriae D. sex pili E. inclusion body \\
\textbf{Answer:} Let’s think step by step. Nosocomial infections are infections acquired in hospitals, with biofilm formation being a major contributing factor. Biofilms are complex structures formed by bacteria adhering to surfaces and producing extracellular polysaccharides, which protect them and increase antibiotic resistance. Among the options, the capsule, a structure made of extracellular polysaccharides surrounding bacteria, plays a critical role in biofilm formation. It helps bacteria adhere to surfaces, connect with each other, and form protective layers. Therefore, the structure associated with biofilm formation in nosocomial infections is the capsule. Thus, the correct answer is (A). \\
\bottomrule
\end{tabular}
}

    \caption{
KormedMCQA Pharmacist CoT Prompt (Translated English Version)
    }
    \label{appendix/table8}
    \end{table}
\newpage

    \begin{table}[!h]
    \centering
    \setlength{\heavyrulewidth}{0.8pt}
    
    \newcolumntype{l}{>{\raggedright\arraybackslash}p{\columnwidth}}
    \small

    \newcommand{\mytable}{\begin{tabular}{l}
\toprule
\textbf{질문:} 국소의치에서 하중부담을 고려하여 최종인상을 채득할 때, 왁스 릴리프의 양을 최대로 확보해야 하는 부위는? \\
\textbf{보기:} A. 절치유두 B. 협측 선반 C. 상악 치조정 D. 상악 치조제의 협측 경사면 E. 하악 치조제의 설측 경사면 \\
\textbf{정답:} 차근차근 생각해봅시다. 국소의치에서 하중부담을 고려하여 최종인상을 채득할 때, 왁스 릴리프의 양을 최대로 확보해야 하는 부위를 선택해야 합니다. 왁스 릴리프는 압박이 가해지는 부위를 보호하고 환자의 불편감을 줄이기 위해 필요합니다. 절치유두는 상악에서 특히 압박을 받을 수 있는 부위로, 이곳에 적절한 완충을 제공하는 것이 중요합니다. 따라서, 이 부위에 왁스 릴리프를 최대한으로 확보해야 합니다. 따라서, 정답은 (A) 입니다. \\
\break
\textbf{질문:} 후방연장 의치상의 지지를 얻기 위해 이차인상을 이용하여 개조모형(altered cast)을 만드는 것이 추천되는 부분무치악 분류는? \\
\textbf{보기:} A. 상악 Kennedy Class I B. 하악 Kennedy Class I C. 상악 Kennedy Class II D. 하악 Kennedy Class III E. 하악 Kennedy Class IV \\
\textbf{정답:} 차근차근 생각해봅시다. 후방연장 의치는 후방에 지지되는 치아가 없으므로 의치의 안정성을 위해 지지를 잘 확보하는 것이 중요합니다. 개조모형은 이차인상을 통해 보다 정확한 연조직의 상태를 반영하여 의치를 제작할 수 있도록 도와줍니다. 특히 후방연장이 필요한 Kennedy Class I의 경우 양측 후방에 지지 치아가 없기 때문에 이차인상을 통해 개조모형을 만드는 것이 중요합니다. 따라서, 상악과 하악의 Kennedy Class I에서 이 방법이 추천됩니다. 따라서, 정답은 (B) 입니다. \\
\break
\textbf{질문:} 치주-보철을 위한 보조적 교정치료 시 고정식 교정장치 대신 가철식 교정장치를 사용하는 이유는? \\
\textbf{보기:} A. 발음 및 저작장애가 없다. B. 치아 이동 모멘트의 부여가 쉽다. C. 이동부 치아의 효과적인 이동이 가능하다. D. 치료결과가 환자의 협조에 의해 좌우되지 않는다. E. 치아 이외의 조직에서 고정원(anchorage)을 추가로 얻을 수 있다. \\
\textbf{정답:} 차근차근 생각해봅시다. 치주-보철을 위한 보조적 교정치료 시 가철식 교정장치를 선택하는 이유는 주로 환자의 편의성과 특정 치료 목표에 맞춘 조절 가능성 때문입니다. 고정식 교정장치에 비해 가철식 교정장치는 환자가 직접 탈착이 가능하여 일상생활에서 발음 및 저작장애가 적고, 필요할 때만 장치를 착용할 수 있다는 장점이 있습니다. 또한, 가철식 교정장치는 치아 이외의 조직에서 추가적인 고정원을 얻을 수 있어 교정치료 시 효과적인 결과를 얻는 데 도움을 줍니다. 따라서, 정답은 (E) 입니다. \\
\break
\textbf{질문:} 편측성 완전구순구개열 환자의 치조열 이차골이식술을 시행 하기에 가장 선호되는 시기는? \\
\textbf{보기:} A. 1세경 B. 5세경 C. 10세경 D. 15세경 E. 20세경 \\
\textbf{정답:} 차근차근 생각해봅시다. 치조열 이차골이식술은 구순구개열 환자의 치조골 결손을 보충하기 위해 수행되는 중요한 수술입니다. 이 수술은 영구견치의 맹출 시기에 맞추어 시행되는 것이 가장 효과적입니다. 왜냐하면 이 시기에 치조골을 이식하면 견치의 정상적인 맹출을 도울 수 있기 때문입니다. 일반적으로 견치는 9세에서 11세 사이에 맹출하므로, 이식술은 이보다 조금 이른 10세경에 시행하는 것이 이상적입니다. 따라서, 정답은 (C) 입니다. \\
\break
\textbf{질문:} 상악 무치악 구강에서 저작압을 가장 많이 지지할 수 있는 부위는? \\
\textbf{보기:} A. 구개융기 B. 정중구개봉선 C. 경구개 후방부위 D. 경구개의 평탄한 부분 E. 구상절흔(hamular notch) \\
\textbf{정답:} 차근차근 생각해봅시다. 상악 무치악 구강에서 저작압을 가장 많이 지지할 수 있는 부위를 선택해야 합니다. 경구개 후방부위는 해부학적으로 두꺼운 골 구조와 함께 저작압을 잘 견딜 수 있는 조직을 포함하고 있습니다. 이는 상악 무치악에서 저작압의 분산 및 지지를 제공하는 주요 부위로 평가됩니다. 따라서, 정답은 (C) 입니다. \\
\bottomrule
\end{tabular}
}

    \caption{
KormedMCQA Dentist CoT Prompt (Original Korean Version)
    }
    \label{appendix/table9}
    \end{table}
\newpage

    \begin{table}[!h]
    \centering
    \setlength{\heavyrulewidth}{0.8pt}
    
    \newcolumntype{l}{>{\raggedright\arraybackslash}p{\columnwidth}}
    \small

    \newcommand{\mytable}{\begin{tabular}{l}
\toprule
\textbf{Question:} When taking the final impression considering load-bearing in removable partial dentures, which area should have the maximum amount of wax relief? \\
\textbf{Options:} A. Incisive papilla B. Buccal shelf C. Maxillary residual ridge D. Buccal slope of the maxillary ridge E. Lingual slope of the mandibular ridge \\
\textbf{Answer:} Let’s think step by step. When taking the final impression for removable partial dentures, the area requiring the maximum wax relief should be chosen carefully to protect against excessive pressure and reduce discomfort for the patient. The incisive papilla is particularly susceptible to pressure in the maxilla, so providing adequate cushioning in this area is critical. Thus, the correct answer is (A). \\
\break
\textbf{Question:} For obtaining support in distal extension denture bases, which type of partially edentulous classification is recommended for creating an altered cast using secondary impressions? \\
\textbf{Options:} A. Maxillary Kennedy Class I B. Mandibular Kennedy Class I C. Maxillary Kennedy Class II D. Mandibular Kennedy Class III E. Mandibular Kennedy Class IV \\
\textbf{Answer:} Let’s think step by step. Distal extension dentures lack posterior supporting teeth, so ensuring proper support is crucial for stability. Altered casts using secondary impressions help reflect the soft tissue's accurate condition for denture fabrication. This method is particularly important in Kennedy Class I cases, where there are no posterior abutment teeth bilaterally. Therefore, it is recommended for both maxillary and mandibular Kennedy Class I cases. Thus, the correct answer is (B). \\
\break
\textbf{Question:} What is the primary reason for choosing removable orthodontic appliances instead of fixed appliances in adjunctive orthodontic treatment for periodontal-prosthetic cases? \\
\textbf{Options:} A. No interference with speech or mastication B. Easy to apply tooth-moving moments C. Effective movement of specific teeth D. Not dependent on patient cooperation E. Additional anchorage from non-dental structures \\
\textbf{Answer:} Let’s think step by step. Removable orthodontic appliances are often chosen for their convenience and adjustability to achieve specific treatment goals. Compared to fixed appliances, removable appliances allow patients to remove them, reducing interference with speech and mastication. Additionally, removable appliances can provide anchorage from non-dental structures, facilitating effective treatment outcomes. Thus, the correct answer is (E). \\
\break
\textbf{Question:} What is the most preferred timing for secondary alveolar bone grafting in patients with unilateral complete cleft lip and palate? \\
\textbf{Options:} A. Around 1 year B. Around 5 years C. Around 10 years D. Around 15 years E. Around 20 years \\
\textbf{Answer:} Let’s think step by step. Secondary alveolar bone grafting is a crucial procedure to address alveolar bone defects in cleft lip and palate patients. This surgery is most effective when performed around the eruption of the permanent canine, as it facilitates normal eruption and alignment. Typically, the canine erupts between ages 9 and 11, making around 10 years of age the optimal time for the grafting. Thus, the correct answer is (C). \\
\break
\textbf{Question:} In the edentulous maxilla, which area can bear the highest masticatory pressure? \\
\textbf{Options:} A. Palatine torus B. Median palatal suture C. Posterior hard palate D. Flat portion of the hard palate E. Hamular notch \\
\textbf{Answer:} Let’s think step by step. The area capable of bearing the highest masticatory pressure in the edentulous maxilla must be selected. The posterior hard palate contains thick bony structures and supportive tissues that efficiently distribute and withstand masticatory forces. It is considered the primary area for supporting masticatory pressure in the edentulous maxilla. Thus, the correct answer is (C). \\
\bottomrule
\end{tabular}
}

    \caption{
KormedMCQA Dentist CoT Prompt (Translated English Version)
    }
    \label{appendix/table10}
    \end{table}
\newpage

\section{Results} \label{appendix:results}

    \begin{table}[!h]
    \centering
    \setlength{\heavyrulewidth}{0.80pt}
    \resizebox{0.95\columnwidth}{!}{
        \newcommand{\mytable}{\begin{tabular}{llcccccccccc}
\toprule
\multicolumn{1}{l}{\textbf{Model}} & \multicolumn{1}{l}{\textbf{Citation}} & \multicolumn{1}{c}{\textbf{Doctor}} & \multicolumn{1}{c}{\textbf{Nurse}} & \multicolumn{1}{c}{\textbf{Pharm}} & \multicolumn{1}{c}{\textbf{Dentist}} & \multicolumn{1}{c}{\textbf{Avg}} & \multicolumn{1}{c}{\textbf{MedQA}} & \multicolumn{1}{c}{\textbf{IFEval}} & \multicolumn{1}{c}{\textbf{BBH}} & \multicolumn{1}{c}{\textbf{GPQA}} & \multicolumn{1}{c}{\textbf{MMLU-PRO}} \\ 
\midrule
\multicolumn{12}{c}{\textbf{Human Performance}}  \\ \midrule
Examinees Average&&79.66&79.90&71.86&78.23&77.05&-&-&-&-&- \\
\midrule 

\multicolumn{12}{c}{\textbf{Proprietary Models}}                                                                                                                                       \\ \midrule
o1-preview-2024-09-12 & \citet{openai2024b} & 92.41 & 95.10 & 94.24 & 88.66 & \textbf{92.72} & 94.19 &-&-&-&- \\
o1-mini-2024-09-12 & \citet{openai2024b} & 88.28 & 85.88 & 87.68 & 69.91 & 82.45 & 90.18 &-&-&-&- \\
gpt-4o-2024-08-06 & \citet{openai2024a} & 85.98 & 91.46 & 85.99 & 78.67 & 85.61 & 86.96 &-&-&-&- \\
gpt-4o-mini-2024-07-18 & \citet{openai2024a} & 63.22 & 83.14 & 71.30 & 59.06 & 70.29 & 72.90 &-&-&-&- \\
gpt-4-turbo-2024-04-09 & \citet{achiam2023gpt} & 76.32 & 86.67 & 81.02 & 66.71 & 78.13 & 76.43 &-&-&-&- \\
gpt-4-0125-preview & \citet{achiam2023gpt} &75.17 & 86.90 & 81.92 & 66.46 & 78.23 & 78.24 &-&-&-&- \\
gpt-3.5-turbo-0125 & \citet{brown2020language} & 31.26 & 52.85 & 46.55 & 32.06 & 42.27 & 78.55 &-&-&-&- \\
claude-3-5-sonnet-20241022&\citet{anthropic2024b}&88.05&92.14&91.41&74.23&86.51 & 82.95 &-&-&-&- \\
claude-3-5-haiku-20241022&\citet{anthropic2024b}&69.66&81.89&58.42&62.27&67.93 & 69.21 &-&-&-&- \\
claude-3-opus-20240229&\citet{anthropic2024a}&74.25&87.24&74.35&69.30&76.74 & 76.12 &-&-&-&- \\
claude-3-sonnet-20240229&\citet{anthropic2024a}&66.90&78.13&63.73&43.25&62.87 & 65.91 &-&-&-&- \\
gemini-1.5-pro&\citet{team2024gemini}&79.31&88.27&80.34&70.28&79.79 & 76.67 &-&-&-&- \\
gemini-1.5-flash&\citet{team2024gemini}&68.51&82.46&74.92&59.68&72.09 & 67.09 &-&-&-&- \\
\midrule

\multicolumn{12}{c}{\textbf{Multilingual Models}}                                                                                                                    \\ \midrule
Llama-3.1-70B-Instruct&\citet{dubey2024llama}&72.18&82.23&77.74&63.13&74.31&79.42&86.69&55.93&14.21&47.88 \\
Llama-3.1-8B-Instruct&\citet{dubey2024llama}&40.69&59.57&55.25&42.05&50.85&63.32&78.56&29.89&2.35&30.68 \\
Llama-3-70B-Instruct&\citet{dubey2024llama}&66.67&81.21&73.90&63.01&72.05&75.88&80.99&50.19	&4.92	&46.74 \\
Llama-3-8B-Instruct&\citet{dubey2024llama}&38.16&55.92&48.14&41.68&47.23&62.37&74.08&	28.24&	1.23&	29.6 \\
Llama-2-70b-chat-hf&\citet{touvron2023llama}&28.51&45.67&38.08&31.69&37.19&50.75&49.58	&4.61	&1.9&	15.92 \\
Llama-2-13b-chat-hf&\citet{touvron2023llama}&27.36&36.67&31.41&30.83&32.20&42.81&39.85&	7.16&	0&	10.26 \\
Llama-2-7b-chat-hf&\citet{touvron2023llama}&21.61&27.45&25.42&25.77&25.56&38.81&39.86	&4.46&	0.45	&7.64 \\
Qwen2.5-72B-Instruct&\citet{yang2024qwen2}&77.24&87.47&82.71&66.21&\textbf{78.86}&77.93&86.38&61.87&16.67&51.40 \\
Qwen2.5-7B-Instruct&\citet{yang2024qwen2}&51.49&27.33&63.39&39.58&44.73&61.19&75.85&34.89&5.48&36.52 \\
Qwen2.5-1.5B-Instruct&\citet{yang2024qwen2}&33.33&47.72&35.48&32.43&37.92&45.56&44.76&19.81&0.78&19.99 \\
Qwen2-72B-Instruct&\citet{yang2024qwen2}&74.71&86.90&82.03&68.68&78.79&75.41&79.89&57.48&16.33&48.92 \\
Qwen2-7B-Instruct&\citet{yang2024qwen2}&46.21&66.86&57.85&44.27&55.14&55.30&56.79&37.81&6.38&31.64 \\
Qwen2-1.5B-Instruct&\citet{yang2024qwen2}&28.97&44.65&32.54&27.87&34.30&39.67&33.71	&13.7&1.57&16.68 \\
Mistral-Large-Instruct-2411&\citet{mistral2024}&71.26&82.46&78.64&62.89&74.44&79.81&84.01&52.74&24.94&50.69 \\
Mistral-Large-Instruct-2407&\citet{mistral2024}&71.49&83.83&80.45&62.89&75.41&79.65&-&-&-&- \\
Mistral-Nemo-Instruct-2407&\citet{mistral2024}&42.30&65.83&58.31&43.03&54.07&60.88&63.80&29.68&5.37&27.97 \\
Mixtral-8x22B-Instruct-v0.1&\citet{jiang2024mixtral}&48.28&69.36&66.44&47.84&59.65&69.91&71.84&44.11&16.44&38.7 \\
Mixtral-8x7B-Instruct-v0.1&\citet{jiang2024mixtral}&36.55&56.95&54.80&38.84&48.49&61.19&55.99&29.74&7.05&29.91 \\
Mistral-7B-Instruct-v0.3&\citet{jiang2023mistral}&29.66&42.37&41.13&31.20&37.16&51.37 &54.65&25.57&3.91&23.06\\
gemma-2-27b-it&\citet{team2024gemmab}&62.30&78.36&72.20&58.57&68.89&67.79&79.78&49.27&16.67&38.35 \\
gemma-2-9b-it&\citet{team2024gemmab}&20.69&61.73&43.95&42.42&45.36&62.53&74.36&42.14&14.77&31.95 \\
gemma-2-2b-it&\citet{team2024gemmab}&13.33&39.41&27.80&28.11&29.18&42.18&56.68&17.98&3.24&17.22 \\
gemma-7b-it&\citet{team2024gemmaa}&28.89&41.80&34.35&29.96&34.70&39.67&38.68&11.88&4.59&7.72 \\
gemma-1.1-2b-it&\citet{team2024gemmaa}&22.30&25.17&22.60&22.32&23.23&28.99&30.67&5.86&2.57&5.37 \\
Yi-1.5-34B-Chat&\citet{young2024yi}&36.09&51.03&48.81&38.84&44.93&63.55&60.67&44.26&15.32&39.12 \\
Yi-1.5-9B-Chat&\citet{young2024yi}&31.26&44.99&38.98&34.77&38.48&52.24&60.46&36.95&11.3&33.06 \\
Yi-34B-Chat&\citet{young2024yi}&33.33&53.30&50.62&38.72&45.70&63.63&46.99&37.62&11.74&34.37 \\
Yi-6B-Chat&\citet{young2024yi}&28.28&33.49&37.85&29.59&32.97&48.70&33.95&17&5.93&22.9 \\
Phi-3.5-mini-instruct&\citet{abdin2024phi}&34.25&49.66&44.29&37.48&42.57&48.78&57.75&36.75&11.97&32.91 \\
Phi-3.5-MoE-instruct&\citet{abdin2024phi}&52.64&76.08&67.23&52.40&63.71&70.52&69.25&48.77&14.09&40.64 \\
Phi-3-mini-4k-instruct&\citet{abdin2024phi}&29.43&33.37&32.09&27.13&30.74&59.07&54.77&36.56&10.96&33.58 \\
Phi-3-medium-4k-instruct&\citet{abdin2024phi}&31.03&46.47&46.21&35.39&41.18&69.76&64.23&49.38&11.52&40.84 \\
SOLAR-10.7B-Instruct&\citet{kim2023solar}&41.84&58.54&52.77&37.85&48.85&54.52&47.37&31.87&7.83&23.76 \\
\midrule
\multicolumn{12}{c}{\textbf{Korean Specific Models}}                                                                                                                  \\ \midrule
EEVE-Korean-Instruct-10.8B-v1.0&\citet{kim2024efficient}&48.97&66.17&58.19&42.42&54.94&53.50&-&-&-&- \\
EXAONE-3.0-7.8B-Instruct&\citet{an2024exaone}&48.28&67.65&56.61&45.87&55.73&50.59&71.93&17.98&2.13&28.63 \\
KULLM3 & \citet{kullm} & 39.54 & 60.02 & 52.09 & 38.35 & 48.89 & 52.47 & - & - & - & - \\
\midrule
\multicolumn{12}{c}{\textbf{Medical Continual Pretrained \& Finetuned Models}}                                                                                                                      \\ \midrule
Meditron3-70B&\citet{chen2023meditron}&70.57&82.69&75.82&62.64&73.51&79.65&-&-&-&- \\
Meditron3-8B&\citet{chen2023meditron}&38.39&58.20&54.92&39.83&49.42&61.98&-&-&-&- \\
meditron-70b&\citet{chen2023meditron}&42.07&56.61&48.36&37.98&47.06&60.02&-&-&-&- \\
meditron-7b&\citet{chen2023meditron}&24.37&26.20&27.12&24.04&25.62&37.63&-&-&-&- \\
llama-3-meerkat-70b-v1.0&\citet{kim2024small}&69.20&80.75&75.59&60.06&71.99&77.85&-&-&-&- \\
llama-3-meerkat-8b-v1.0&\citet{kim2024small}&38.62&56.61&45.65&40.57&46.46&62.53&-&-&-&- \\
meerkat-7b-v1.0&\citet{kim2024small}&35.17&45.44&44.18&33.66&40.41&61.51&-&-&-&- \\
ClinicalCamel-70B&\citet{toma2023clinical}&43.35&65.15&53.90&42.54&52.61&59.62&-&-&-&- \\
Palmyra-Med-70B&\citet{kamble2023palmyra}&66.67&79.27&73.45&61.65&70.99&77.93&-&-&-&- \\
BioMistral-7B&\citet{labrak2024biomistral}&27.82&37.24&36.16&26.63&32.70&44.38 &-&-&-&-
\\
     \bottomrule
\end{tabular}}
            
    }
    \caption{
Detailed performance results of LLMs on the KorMedMCQA, MedQA, IFEval, BBH, GPQA, and MMLU-Pro.
    }
    \label{appendix:score}
    \end{table}

\newpage
\section{CoT Error Analysis} \label{appendix_cot}


In this section, we present an error analysis of three leading models—GPT-4o, Claude-3.5-sonnet, and Gemini-1.5-pro—on the KorMedMCQA dataset. 
Our analysis identifies specific shortcomings and offers valuable insights to guide future improvements in training strategies. 
We examined 200 randomly selected questions, equally drawn from four professional groups: doctors, nurses, pharmacists, and dentists. 
Each question was incorrectly answered by at least two of the three models. 
Healthcare experts reviewed and classified these errors, with examples in Tables \ref{appendix/table13} to \ref{appendix/table12}.
Examples include the original questions, the models' answers, and the expert explanations. 
For global accessibility, translated versions are also included.

\paragraph{Lack of Clinical Knowledge (36\%)}
The most common error was a lack of specialized medical knowledge in certain areas, leading to incorrect answers. 
For instance, in Table \ref{appendix/table13}, the question asked which anatomical structure determines the fit of a lower front denture. 
The correct answer is the mentalis muscle, which plays a crucial role in the stability of such dentures. 
However, two of the models incorrectly chose the orbicularis oris muscle. 
This suggests that the models lacked specific knowledge in dental anatomy and prosthetics, relying on general understanding or common associations rather than the specialized information required to answer correctly.

\paragraph{Lack of Knowledge in Korean Healthcare Law and Regulations (34\%)}  
The models also demonstrated a lack of knowledge in Korean healthcare laws and regulations. 
This limitation likely stemmed from training data that often lacked localized legislative context. 
For instance, they incorrectly assigned financial responsibility in emergency treatment scenarios (see Table \ref{appendix/table11}). 
These errors arose from overgeneralizing international standards or making incorrect assumptions about the Korean healthcare system, leading to flawed reasoning.

\paragraph{Clinical Reasoning and Diagnostic Errors (24\%)} 
Another category of errors stemmed from the models' misapplication of clinical knowledge during diagnostic reasoning. 
Even though the models possessed the necessary medical information, they sometimes failed to integrate all clinical data properly, leading to incorrect conclusions. 
For example, in Table \ref{appendix/table12}, the models were presented with a patient exhibiting certain symptoms after surgery. 
While they correctly considered a possible cause and suggested treatment, they overlooked key clinical signs that pointed to a different issue. 
This indicates that the models focused on common complications without fully analyzing all of the patient's data.

\paragraph{Other Errors}
Errors not defined in the categories above included Answer Annotation Errors (2\%) and Question Annotation Errors (4\%). 
Answer Annotation Errors occurred when the answers provided by the Korea Health Personnel Licensing Examination Institute were incorrectly annotated. 
When such errors are identified, the institute uploads corrections, which we plan to incorporate in the next version of our dataset. 
The second type, Question Annotation Errors, involved cases where we failed to exclude questions requiring multimedia content. 
For example, a question might ask, "What is the arrhythmia shown in the following ECG?" without providing the ECG image. 
Although such cases were rare, we intend to exclude them in the next version update.

\newpage

    \begin{table}[!h]
    \centering
    \setlength{\heavyrulewidth}{0.5pt}
    
    \newcolumntype{l}{>{\raggedright\arraybackslash}p{\columnwidth}}
    \small

    \newcommand{\mytable}{\begin{tabular}{l}
\toprule
\textbf{질문:} 하악 의치상의 전방부 순측 변연 길이를 결정하는 해부학적 구조물은? \\
\textbf{보기:} A. 볼근(buccinator m.) B. 턱끝근(mentalis m.) C. 턱목뿔근(mylohyoid m.) D. 턱끝혀근(genioglossus m.) E. 입둘레근(orbicularis oris m.)\\
\textbf{정답:} B \\
\textbf{GPT:} 차근차근 생각해봅시다. 하악 의치상의 전방부 순측 변연 길이는 의치가 구강 내에서 안정적으로 위치할 수 있도록 해부학적 구조물에 의해 결정됩니다. 이 부위의 변연 길이는 주로 입술과 관련된 근육에 의해 영향을 받습니다. 입둘레근(orbicularis oris m.)은 입술을 둘러싸고 있는 근육으로, 하악 의치상의 전방부 순측 변연 길이에 직접적인 영향을 미칩니다. 이 근육은 입술의 움직임과 관련이 있으며, 의치의 변연이 이 근육의 기능을 방해하지 않도록 조정되어야 합니다. 따라서, 정답은 (E) 입니다. \\
\textbf{Claude:} 차근차근 생각해봅시다. 하악 의치상의 전방부 순측 변연 길이를 결정할 때는 해당 부위의 근육 부착과 움직임을 고려해야 합니다. 전방부 순측 변연은 입술 부위에 해당하며, 이 부위의 근육 활동은 의치의 안정성에 직접적인 영향을 미칩니다. 이 부위에서 가장 중요한 근육은 턱끝근(mentalis muscle)입니다. 턱끝근은: 하악 전치부 아래쪽에 위치, 입술을 아래로 당기고 앞으로 내미는 기능, 의치상의 변연 길이를 제한하는 주요 요소, 과도한 변연 연장 시 의치의 탈락을 유발할 수 있음. 다른 근육들(볼근, 턱목뿔근, 턱끝혀근, 입둘레근)은 하악 의치상의 다른 부위에 영향을 미치지만, 전방부 순측 변연 길이를 직접적으로 결정하지는 않습니다.
따라서, 정답은 (B) 입니다. \\
\textbf{Gemini:} 차근차근 생각해봅시다. 하악 의치상의 전방부 순측 변연 길이는 의치의 안정성과 심미성에 중요한 영향을 미칩니다. 이 변연 길이는 주변 근육들의 움직임을 고려하여 결정해야 하는데, 특히 입둘레근(orbicularis oris m.)은 입술의 모양을 형성하고 움직임을 담당하는 근육으로, 하악 의치상의 전방부 순측 변연 길이를 결정하는 데 가장 중요한 해부학적 구조물입니다. 따라서, 정답은 (E) 입니다.\\
\textbf{설명: } 모델들은 하악 의치상의 전방부 순측 변연 길이에 가장 큰 영향을 미치는 근육이 턱끝근(mentalis m.)임을 인지하지 못하고, 대신 입둘레근(orbicularis oris m.)의 역할을 과대평가하여 오답을 선택했습니다. 이는 치과 해부학과 의치 제작에 관한 전문 지식의 부족에서 비롯된 오류입니다.\\
\break
\textbf{Question:} Which anatomical structure determines the labial margin length of the anterior mandibular denture? \\ 
\textbf{Options:} A. Buccinator m. B. Mentalis m. C. Mylohyoid m. D. Genioglossus m. E. Orbicularis oris m. \\ \textbf{Answer:} B \\
\textbf{GPT:} Let’s think step by step. The labial margin length of the anterior mandibular denture is determined by anatomical structures to ensure the denture remains stable in the oral cavity. This margin length is primarily influenced by muscles associated with the lips. The orbicularis oris m. encircles the lips and directly impacts the labial margin length of the anterior mandibular denture. This muscle is involved in lip movement, and the denture margin should be adjusted to avoid interfering with its function. Therefore, the correct answer is (E).  \\
\textbf{Claude:} Let’s think step by step. When determining the labial margin length of the anterior mandibular denture, it is crucial to consider muscle attachments and movements in that region. The labial margin corresponds to the area near the lips, where muscle activity directly impacts denture stability. The most critical muscle in this region is the mentalis muscle. The mentalis m. is located below the mandibular anterior teeth and plays a role in pulling the lower lip downward and outward. It is a key factor in limiting the denture margin length, as overextension of the margin can lead to denture dislodgement. Other muscles (buccinator, mylohyoid, genioglossus, orbicularis oris) affect other parts of the mandibular denture but do not directly determine the labial margin length of the anterior denture. Therefore, the correct answer is (B). \\
\textbf{Gemini:} Let’s think step by step. The labial margin length of the anterior mandibular denture significantly influences its stability and aesthetics. This margin length must be determined with consideration of surrounding muscle movements. Specifically, the orbicularis oris m. is the muscle that shapes and controls lip movement, making it the most important anatomical structure for determining the labial margin length of the anterior mandibular denture. Therefore, the correct answer is (E).  \\
\textbf{Explanation:} The models failed to recognize that the muscle most influential in determining the labial margin length of the anterior mandibular denture is the mentalis m. Instead, they overemphasized the role of the orbicularis oris m., leading to incorrect answers. This error stems from a lack of specialized knowledge in dental anatomy and denture fabrication. \\
\bottomrule
\end{tabular}
}

    \caption{
Error Analysis Case 1: Lack of Medical / Clinical Knowledge
    }
    \label{appendix/table13}
    \end{table}

\newpage

    \begin{table}[!h]
    \centering
    \setlength{\heavyrulewidth}{0.8pt}
    
    \newcolumntype{l}{>{\raggedright\arraybackslash}p{\columnwidth}}
    \small

    \newcommand{\mytable}{\begin{tabular}{l}
\toprule
\textbf{질문:} 40세 남자가 군지역의 지역응급의료기관에서 단순골절로 응급진료를 받았다. 환자는 무연고자이고 진료비를 내지 않고 도주하였다. 이때 의료기관을 운용하는 자가 받지 못한 진료비 본인부담금을 대신 지급하여 줄 것을 청구할 수 있는 대상은? \\
\textbf{보기:} A. 군수 B. 대한적십자사 회장 C. 중앙응급의료센터장 D. 건강보험심사평가원장 E. 국민건강보험공단 이사장 \\
\textbf{정답:} D \\
\textbf{GPT:} 차근차근 생각해봅시다. 40세 남자가 군지역의 지역응급의료기관에서 단순골절로 응급진료를 받았으나, 무연고자이고 진료비를 내지 않고 도주하였습니다. 이 경우, 의료기관은 진료비 본인부담금을 대신 지급받기 위해 특정 기관에 청구할 수 있습니다. 응급의료에 관한 법률에 따르면, 응급의료기관은 응급환자에 대한 진료비 중 본인부담금을 환자에게 받을 수 없는 경우, 국민건강보험공단에 그 금액을 청구할 수 있습니다. 국민건강보험공단은 응급의료비 대지급 제도를 통해 이러한 경우에 의료기관에 본인부담금을 지급하고, 이후 환자에게 이를 청구하는 절차를 밟습니다. 따라서, 정답은 (E) 국민건강보험공단 이사장입니다. \\
\textbf{Claude:} 차근차근 생각해봅시다.
1. 응급의료에 관한 법률에 따르면, 응급환자가 진료비를 지불할 능력이 없거나 무연고자인 경우에도 응급의료기관은 응급진료를 거부할 수 없습니다.
2. 이러한 경우 발생하는 미수금에 대해서는 지방자치단체가 보상해주도록 법적으로 규정되어 있습니다.
3. 특히 군지역의 경우, 해당 지역의 군수가 미수금 보상의 책임을 지게 됩니다.
4. 응급의료에 관한 법률 제22조에 따르면, 응급환자의 진료비를 받지 못한 응급의료기관은 관할 지방자치단체의 장에게 대신 지급해줄 것을 청구할 수 있습니다.
5. 이 사례에서는 군지역의 지역응급의료기관이므로, 해당 지역의 군수에게 청구해야 합니다.
따라서, 정답은 (A) 입니다. \\
\textbf{Gemini:} 차근차근 생각해봅시다. 문제는 군지역 응급의료기관에서 무연고자 환자가 진료비를 지불하지 않고 도주한 경우, 의료기관 운영자가 진료비를 대신 받을 수 있는 곳을 묻고 있습니다. 응급의료에 관한 법률 제35조(응급진료비용 미수금 대지급)에 따르면, 응급의료기관의 장은 응급환자에게 발생한 응급진료비용 중 본인부담금을 받지 못한 경우 그 미수금에 대하여 국민건강보험공단에 대신 지급하여 줄 것을 청구할 수 있습니다. 따라서, 정답은 (E) 입니다.\\
\textbf{설명: } 의료기관은 진료비 본인부담금을 대신 지급받기 위해 특정 기관에 청구할 수 있습니다. 응급의료에 관한 법률에 따르면, 응급의료기관은 응급환자에 대한 진료비 중 본인부담금을 환자에게 받을 수 없는 경우, 심사평가원에 그 금액을 청구할 수 있습니다. 그러므로 심사평가원이 답입니다. 모든 LLM들은 응급 환자 미수금을 대납 받을 수 있음은 파악하였지만, 해당 미수금 청구 대상이 정확히 어떤 기관인지 파악하지 못했습니다. \\
\break
\textbf{Question:} A 40-year-old man received emergency treatment for a simple fracture at a regional emergency medical institution in a military area. The patient, who is unaccompanied and has no personal connections, fled without paying the treatment fee. In this case, who can the operator of the medical institution claim the unpaid co-payment for medical expenses from? \\ \textbf{Options:} A. County Governor B. President of the Korean Red Cross C. Director of the Central Emergency Medical Center D. Director of the Health Insurance Review and Assessment Service E. President of the National Health Insurance Corporation \\
\textbf{Answer:} D \\
\textbf{GPT:} Let’s think this step by step. A 40-year-old man received emergency treatment for a simple fracture at a regional emergency medical institution in a military area. The patient, being unaccompanied and fleeing without paying the fee, leaves the medical institution to seek compensation for the unpaid co-payment. According to the Emergency Medical Service Act, if an emergency medical institution cannot collect the co-payment from the patient, they can claim the amount from the National Health Insurance Corporation. The National Health Insurance Corporation operates a system to cover unpaid emergency medical expenses and later recovers the amount from the patient. Therefore, the answer is (E) President of the National Health Insurance Corporation. \\
\textbf{Claude:} Let’s think step by step. Under the Emergency Medical Service Act, emergency medical institutions cannot refuse emergency treatment for patients unable to pay or unaccompanied individuals.
Unpaid expenses in such cases are legally compensated by the local government.
Specifically, in military areas, the head of the county assumes responsibility for such compensation.
According to Article 22 of the Emergency Medical Service Act, an emergency medical institution that cannot collect payment from an emergency patient can claim the amount from the head of the local government.
Since this case involves a regional emergency medical institution in a military area, the claim should be made to the county governor. Thus, the answer is (A). \\
\textbf{Gemini:} Let’s think step by step. The question asks which institution the operator of the medical institution can claim unpaid treatment fees from in the case of an unaccompanied patient who fled without paying. According to Article 35 of the Emergency Medical Service Act (Payment for Unpaid Emergency Treatment Costs), if an emergency medical institution cannot collect co-payment fees from an emergency patient, it can claim the unpaid amount from the National Health Insurance Corporation. Therefore, the answer is (E). \\
\textbf{Explanation:} The medical institution can claim unpaid co-payment fees from a specific organization. According to the Emergency Medical Service Act, an emergency medical institution unable to collect co-payments for emergency treatment costs can claim the amount from the Health Insurance Review and Assessment Service. Thus, the correct answer is the Health Insurance Review and Assessment Service. All LLMs identified the provision for unpaid emergency medical costs but failed to pinpoint the precise institution responsible for reimbursement. \\
\bottomrule
\end{tabular}
}

    \caption{
Error Analysis Case 2: Lack of Knowledge in Korean Healthcare Law / Regulations
    }
    \label{appendix/table11}
    \end{table}

\newpage

    \begin{table}[!h]
    \centering
    \setlength{\heavyrulewidth}{0.5pt}
    
    \newcolumntype{l}{>{\raggedright\arraybackslash}p{\columnwidth}}
    \small

    \newcommand{\mytable}{\begin{tabular}{l}
\toprule
\textbf{질문:} 45세 여자가 입 주위와 손발이 저리다고 한다. 3시간 전 갑상샘암으로 오른쪽 갑상샘절제를 받았다. 혈압 119/78 mmHg, 맥박 89회/분, 호흡 29회/분, 체온 36.7℃이다. 목은 붓지 않았다. 가슴 청진에서 호흡음은 정상으로 들린다. 혈액검사 결과는 다음과 같다. 조치는?혈액: Na +/K+/Cl- 143/3.8/101 meq/L Ca/P 8.6/4.1 mg/dL, 부갑상샘호르몬(iPTH) 19 pg/mL (참고치, 8～51) 갑상샘자극호르몬 3.21 mIU/L (참고치, 0.34～4.25) 유리 T4 1.3 ng/dL (참고치, 0.8～1.7) 동맥혈기체(대기호흡) : pH 7.49, PaCO2 29 mmHg PaO2 97 mmHg, HCO3- 22.2 meq/L \\
\textbf{보기:} A. 경과관찰 B. 산소흡입 C. 재호흡주머니 적용 D. 레보타이록신 투여 E. 10\% 글루콘산칼슘 정맥주사 \\
\textbf{정답:} C \\
\textbf{GPT:} 차근차근 생각해봅시다. 45세 여자는 갑상샘암으로 인해 오른쪽 갑상샘절제를 받은 후 입 주위와 손발이 저리는 증상을 호소하고 있습니다. 이러한 증상은 저칼슘혈증과 관련이 있을 수 있습니다. 혈액검사 결과를 보면, 혈청 칼슘 수치가 8.6 mg/dL로 정상 하한선에 가깝지만, 수술 후 저칼슘혈증의 증상으로 나타날 수 있습니다. 또한, 부갑상샘호르몬(iPTH) 수치가 정상 범위 내에 있지만, 수술 후 부갑상샘 기능 저하로 인해 상대적으로 낮을 수 있습니다. 동맥혈기체 분석에서 pH가 7.49로 알칼리증을 보이고 있으며, 이는 호흡성 알칼리증을 시사합니다. 그러나, 저칼슘혈증으로 인한 신경과민 증상이 더 우선적으로 해결되어야 합니다. 이러한 경우, 10\% 글루콘산칼슘 정맥주사를 통해 칼슘을 보충하는 것이 적절한 조치입니다. 따라서, 정답은 (E) 10\% 글루콘산칼슘 정맥주사입니다. \\
\textbf{Claude:} 차근차근 생각해봅시다.
1. 임상적 판단:
- 갑상샘 수술 후 발생한 저린 증상은 수술 중 부갑상샘 손상이나 일시적인 기능저하로 인한 저칼슘혈증을 시사
- 호흡이 빨라진 것은 저칼슘혈증으로 인한 보상기전일 수 있음
- 입 주위 저림은 Chvostek 징후일 수 있으며, 이는 저칼슘혈증의 전형적인 증상
2. 필요한 조치:
- 급성 저칼슘혈증에 대한 즉각적인 교정이 필요
- 10\% 글루콘산칼슘의 정맥 투여가 가장 적절한 치료
따라서, 정답은 (E) 입니다. \\
\textbf{Gemini:} 차근차근 생각해봅시다. 45세 여자 환자는 갑상샘암으로 오른쪽 갑상샘절제술을 받은 후 입 주위와 손발 저림을 호소합니다. 이는 부갑상샘 기능저하증을 시사하는 증상입니다. 혈액 검사 결과를 보면 칼슘 수치는 8.6 mg/dL로 정상 범위보다 약간 낮고, 부갑상샘호르몬(iPTH) 수치는 19 pg/mL로 정상 범위 안에 있지만, 갑상샘절제술 후에는 일반적으로 iPTH 수치가 상승해야 합니다. 이는 수술 중 부갑상샘이 손상되었을 가능성을 시사하며, 환자의 저칼슘혈증 증상과 함께 부갑상샘 기능저하증을 의심하게 합니다. 따라서 저칼슘혈증 증상을 조절하기 위해 10\% 글루콘산칼슘 정맥주사를 투여하는 것이 적절한 조치입니다. 따라서, 정답은 (E) 입니다.\\
\textbf{설명: } 이 환자는 얼핏 갑상샘 수술 이후 부갑상샘 저하증으로 생각될 수 있습니다. 각 모델은 모두 부갑상샘 기능저하를 의심하고 있습니다. 이에 따라, 각 모델들은 치료로 칼슘 글루코네이트 주사를 추천합니다. 하지만, 일반적으로 칼슘 글루코네이트는 훨씬 더 중증의 저칼슘 혈증에서 사용됩니다.(7.2 mg/dL 이하). 환자는 비교적 경증의 증상만 보이며 칼슘 농도도 Lower boundary 수준입니다. 그에 따라, 굳이 칼슘 글루코네이트 주사까진 필요 없어 보입니다. 또한,환자는 동맥혈 가스검사에서 pH 7.49, PaCO2 29 mmHg로 호흡성 알칼리증을 보이고 분당 29회의 빠른 호흡을 보이고 있습니다. 이를 고려하였을 때 재호흡 주머니를 적용하여 호흡성 알칼리증을 개선하는 것이 필요합니다.  \\
\break
\textbf{Question:} A 45-year-old woman complains of numbness around her mouth and in her hands and feet. She underwent a right thyroid lobectomy for thyroid cancer 3 hours ago. Her vital signs are as follows: blood pressure 119/78 mmHg, pulse rate 89 bpm, respiratory rate 29 breaths/min, and temperature 36.7℃. Her neck is not swollen, and chest auscultation reveals normal breath sounds. Laboratory results are as follows: Blood: Na+/K+/Cl-: 143/3.8/101 mEq/L, Ca/P: 8.6/4.1 mg/dL, parathyroid hormone (iPTH): 19 pg/mL (reference range: 8–51), thyroid-stimulating hormone: 3.21 mIU/L (reference range: 0.34–4.25), free T4: 1.3 ng/dL (reference range: 0.8–1.7). Arterial blood gas (room air): pH 7.49, PaCO2 29 mmHg, PaO2 97 mmHg, HCO3- 22.2 mEq/L. What is the appropriate management? \\ \textbf{Options:} A. Observation B. Oxygen inhalation C. Application of a rebreathing bag D. Administration of levothyroxine E. Intravenous injection of 10\% calcium gluconate \\ \textbf{Answer:} C \\
\textbf{GPT:} Let’s think step by step. A 45-year-old woman complains of tingling around the mouth and in her hands and feet after undergoing a right thyroidectomy due to thyroid cancer. These symptoms could be related to hypocalcemia. From the blood test results, her serum calcium level is 8.6 mg/dL, which is near the lower limit of normal but could manifest as postoperative hypocalcemia. Additionally, her parathyroid hormone (iPTH) levels are within the normal range but may be relatively low due to decreased parathyroid function post-surgery. Arterial blood gas analysis shows a pH of 7.49, indicating alkalosis, which suggests respiratory alkalosis. However, symptoms of neuromuscular irritability due to hypocalcemia should be addressed first. In such cases, administering intravenous 10\% calcium gluconate is an appropriate intervention. Therefore, the correct answer is (E) intravenous 10\% calcium gluconate.  \\
\textbf{Claude:} Let’s think step by step. 3. Clinical judgment: - The tingling symptoms after thyroid surgery suggest hypocalcemia due to potential parathyroid gland damage or temporary dysfunction. - Rapid breathing could be a compensatory mechanism for hypocalcemia. - Tingling around the mouth could indicate Chvostek’s sign, a classic symptom of hypocalcemia. 4. Necessary measures: - Immediate correction of acute hypocalcemia is required. - Intravenous administration of 10\% calcium gluconate is the most appropriate treatment.
Therefore, the correct answer is (E). \\
\textbf{Gemini:} Let’s think step by step. A 45-year-old female patient complains of tingling around the mouth and in her hands and feet following a right thyroidectomy for thyroid cancer. This suggests hypoparathyroidism. Blood test results show a calcium level of 8.6 mg/dL, slightly below normal, and an iPTH level of 19 pg/mL, within the normal range. However, after thyroidectomy, iPTH levels would typically be elevated. This raises the suspicion of parathyroid damage during surgery. Combined with the patient’s hypocalcemic symptoms, hypoparathyroidism is likely. To manage her symptoms, administering intravenous 10\% calcium gluconate is an appropriate measure. Thus, the correct answer is (E).  \\
\textbf{Explanation:} At first glance, this case could be attributed to postoperative hypoparathyroidism following thyroid surgery. All models suspect parathyroid dysfunction and recommend calcium gluconate injections as treatment. However, calcium gluconate is generally used for more severe hypocalcemia (below 7.2 mg/dL). The patient has relatively mild symptoms and a calcium level near the lower boundary of normal. Therefore, intravenous calcium gluconate may not be strictly necessary. Additionally, the patient shows respiratory alkalosis with a pH of 7.49 and PaCO2 of 29 mmHg, along with a rapid respiratory rate of 29 breaths/min. Considering this, applying a rebreathing bag to address respiratory alkalosis would be appropriate.  \\
\bottomrule
\end{tabular}
}

    \caption{
Error Analysis Case 3: Clinical Reasoning / Diagnosis Error
    }
    \label{appendix/table12}
    \end{table}

\newpage
\newpage
\section{Datasheet for Datasets}
\paragraph{A.1\;\;\;\;Motivation}
\begin{itemize}
    \item \textbf{For what purpose was the dataset created?}
    
    The KorMedMCQA dataset was created to evaluate the performance of language models in answering medical multiple-choice questions specifically in the Korean context. It aims to address the need for country-specific medical QA benchmarks that take into account linguistic, clinical, and regulatory differences in healthcare systems. By providing this dataset, the goal is to advance research in healthcare AI tailored for South Korea.
    
    \item \textbf{Who created the dataset (e.g., which team, research group) and on behalf of which entity (e.g., company, institution, organization)?}
    
    The dataset was created by the authors of the paper.

    \item \textbf{Who funded the creation of the dataset? If there is an associated grant, please provide the name of the grantor and the grant name and number.}

    N/A

\end{itemize}

\paragraph{A.2\;\;\;\;Composition}
\begin{itemize}
    \item \textbf{What do the instances that comprise the dataset represent (e.g., documents, photos, people, countries)?}

    The instances represent multiple-choice questions derived from official Korean healthcare professional licensing examinations, including questions for doctors, nurses, pharmacists, and dentists.
    
    \item \textbf{How many instances are there in total (of each type, if appropriate)?}

    The dataset consists of 7,469 multiple-choice questions: Doctors (2,489 questions), Nurses (1,751 questions), Pharmacists (1,817 questions), and Dentists (1,412 questions).

    \item \textbf{Does the dataset contain all possible instances or is it a sample (not necessarily random) of instances from a larger set?}

    The dataset contains all publicly available questions from the Korean healthcare licensing examinations conducted between 2012 and 2024. However, Questions with multiple correct answers were excluded to maintain consistency. Similarly, questions requiring image or video contents were excluded, as these elements are often masked in the original data due to copyright issues.

    \item \textbf{What data does each instance consist of?}

    Each instance includes: A unique identifier (denoting year, session, and question number), the question text, five answer choices, and the index of the correct answer.

    \item \textbf{Is there a label or target associated with each instance?}

    Yes, the correct answer (label) is provided for each instance.

    \item \textbf{Is any information missing from individual instances? If so, please provide a description, explaining why this information is missing (e.g., because it was unavailable). This does not include intentionally removed information, but might include, e.g., redacted text.}

    No information is missing from the instances.
    
    \item \textbf{Are relationships between individual instances made explicit (e.g., users' movie ratings, social network links)?}

    Not Applicable. Each question is independent and does not have explicit relationships with others.

    \item \textbf{Are there recommended data splits (e.g., training, development/validation, testing)?}

    Yes, the dataset is split as follows: Training set (Questions from years before 2021), Development set (Questions from 2021) and Test set (Questions from 2022–2024).

    \item \textbf{Are there any errors, sources of noise, or redundancies in the dataset?}

As noted in Appendix \ref{appendix_cot}, there are some annotation errors in the dataset: Answer annotation errors and Question annotation errors
Answer annotation errors occured when the Korea Health Personnel Licensing Examination Institute provides incorrectly annotated answers. When such errors are identified, the institute issues corrections, which we plan to incorporate into the next version of our dataset.
Question annotation errors arose when questions requiring multimedia content are not excluded. For instance, a question might ask, "What is the arrhythmia shown in the following ECG?" without including the ECG image. While these cases are rare, we aim to exclude them in the next version update.

    \item \textbf{Is the dataset self-contained, or does it link to or otherwise rely on external resources (e.g., websites, tweets, other datasets)?}

    The dataset is self-contained and does not rely on external resources.

    \item \textbf{Does the dataset contain data that might be considered confidential (e.g., data that is protected by legal privilege or by doctor-patient confidentiality, data that includes the content of individuals' non-public communications)?}

    No, the dataset does not contain confidential data. All questions are from publicly released licensing exams.

    \item \textbf{Does the dataset contain data that, if viewed directly, might be offensive, insulting, threatening, or might otherwise cause anxiety?}

    No, the dataset is free of offensive or distressing content.

    \item \textbf{Does the dataset relate to people?}

    Yes, it pertains to professional healthcare licensing exams, which indirectly relate to healthcare professionals.
    
    \item \textbf{Does the dataset identify any subpopulations (e.g., by age, gender)?}

    No, the dataset does not identify any subpopulations.

    \item \textbf{Does the dataset contain data that might be considered sensitive in any way (e.g., data that reveals race or ethnic origins, sexual orientations, religious beliefs, political opinions or union memberships, or locations; financial or health data; biometric or genetic data; forms of government identification, such as social security numbers; criminal history)?}

    No, the original source data has been already de-identified, and no sensitive information is included.

\end{itemize}

\paragraph{A.3\;\;\;\;Collection process}
\begin{itemize}
    \item \textbf{How was the data associated with each instance acquired?}

    The dataset was directly sourced from official medical licensing examination materials released by the Korea Health Personnel Licensing Examination Institute\footnote{\url{https://www.kuksiwon.or.kr}}. These materials were publicly available for academic use. The dataset includes questions from licensing exams for doctors, nurses, pharmacists, and dentists conducted between 2012 and 2024.

    \item \textbf{What mechanisms or procedures were used to collect the data (e.g., hardware apparatuses or sensors, manual human curation, software programs, software APIs)?}

    Data was collected by programmatically crawling publicly available PDF documents from the official Korea Health Personnel Licensing Examination Institute website. These documents were then parsed and organized into a structured dataset. Manual curation was performed to exclude questions with multiple correct answers or those requiring image or video content, ensuring consistency across entries.

    \item \textbf{If the dataset is a sample from a larger set, what was the sampling strategy (e.g., deterministic, probabilistic with specific sampling probabilities)?}

    The dataset comprises all publicly available exam questions from 2012 to 2024, without any sampling. Questions requiring multimedia inputs or having multiple answers were excluded to maintain consistency.

    \item \textbf{Who was involved in the data collection process (e.g., students, crowd workers, contractors) and how were they compensated (e.g., how much were crowd workers paid)?}

    The data collection and curation process were conducted exclusively by the authors of this study. No external contractors or crowd workers were involved.

    \item \textbf{Over what timeframe was the data collected?}

    The data was collected and curated between in 2024. The source materials span medical licensing exams conducted from 2012 to 2024.

    \item \textbf{Were any ethical review processes conducted (e.g., by an institutional review board)?}

    No ethical review process was required.

    \item \textbf{Did you collect the data from the individuals in question directly, or obtain it via third parties or other sources (e.g., websites)?}

    The data was obtained from third-party public sources, specifically the Korea Health Personnel Licensing Examination Institute website.

    \item \textbf{Were the individuals in question notified about the data collection?}

    Not applicable.

    \item \textbf{Did the individuals in question consent to the collection and use of their data?}

    Not applicable.

    \item \textbf{If consent was obtained, were the consenting individuals provided with a mechanism to revoke their consent in the future or for certain uses?}

    Not applicable.

    \item \textbf{Has an analysis of the potential impact of the dataset and its use on data subjects (e.g., a data protection impact analysis) been conducted?}

    The dataset consists of anonymized examination materials and does not contain any personal or sensitive information. Thus, a data protection impact analysis was not deemed necessary.

\end{itemize}

\paragraph{A.4\;\;\;\;Preprocessing/cleaning/labeling}
\begin{itemize}
    \item \textbf{Was any preprocessing/cleaning/labeling of the data done (e.g., discretization or bucketing, tokenization, part-of-speech tagging, SIFT feature extraction, removal of instances, processing of missing values)?}

    Manual curation was performed to exclude questions with multiple correct answers or those requiring image or video content, ensuring consistency across entries.

    \item \textbf{Was the ``raw'' data saved in addition to the preprocess/cleaned/labeled data (e.g., to support unanticipated future uses)?}

    Not applicable.

    \item \textbf{Is the software that was used to preprocess/clean/label the data available?}

    Not applicable.

\end{itemize}

\paragraph{A.5\;\;\;\;Uses}
\begin{itemize}
    \item \textbf{Has the dataset been used for any tasks already?}

    Not applicable.

    \item \textbf{Is there a repository that links to any or all papers or systems that use the dataset?}

    Not applicable.

    \item \textbf{What (other) tasks could the dataset be used for?}

    Beyond its primary use in medical multiple-choice question answering, the dataset could be extended to: 1) Training and evaluating language models in domain-specific QA tasks. 2) Research on cross-lingual transfer of medical knowledge. 3) Studies on reasoning, explanation generation, and contextual understanding in healthcare AI. 4) Educational tools for medical professionals in Korea.

    \item \textbf{Is there anything about the composition of the dataset or the way it was collected and preprocessed/cleaned/labeled that might impact future uses?}

    Not applicable.

    \item \textbf{Are there tasks for which the dataset should not be used?}

    Not applicable.

\end{itemize}

\paragraph{A.6\;\;\;\;Distribution}
\begin{itemize}
    \item \textbf{Will the dataset be distributed to third parties outside of the entity (e.g., company, institution, organization) on behalf of which the dataset was created?}

    No.

    \item \textbf{How will the dataset be distributed?}

    The dataset is publicly available via Huggingface.

    \item \textbf{Will the dataset be distributed under a copyright or other intellectual property (IP) license, and/or under applicable terms of use (ToU)?}

    The dataset is released under MIT license.

    \item \textbf{Have any third parties imposed IP-based or other restrictions on the data associated with the instances?}

    No.

    \item \textbf{Do any export controls or other regulatory restrictions apply to the dataset or to individual instances?}

    No.

\end{itemize}

\paragraph{A.7\;\;\;\;Maintenance}
\begin{itemize}
    \item \textbf{Who will be supporting/hosting/maintaining the dataset?}

    The authors of the paper will be responsible for supporting, hosting, and maintaining the dataset.

    \item \textbf{How can the owner/curator/manager of the dataset be contacted(e.g., email address)?}

    Contact first author : sean0042@kaist.ac.kr

    \item \textbf{Is there an erratum?}

    Currently, there is no erratum.

    \item \textbf{Will the dataset be updated (e.g., to correct labeling erros, add new instances, delete instances)?}

    Yes. Updates will be provided to correct any errors or inconsistencies, and new instances will be added annually to incorporate newly released examination. Detailed explanations of updates will accompany any new versions.

    \item \textbf{If the dataset relates to people, are there applicable limits on the retention of the data associated with the instances (e.g., were the individuals in question told that their data would be retained for a fixed period of time and then deleted)?}

    Not Applicable.

    \item \textbf{Will older versions of the dataset continue to be supported/hosted/maintained?}

    Yes. Older versions of the dataset will be retained and made accessible to ensure reproducibility and comparison with prior work.

    \item \textbf{If others want to extend/augment/build on/contribute to the dataset, is there a mechanism for them to do so?}

    Contact first author : sean0042@kaist.ac.kr

\end{itemize}

\end{document}